\documentclass[10pt,twocolumn,letterpaper]{article}

\usepackage{wacv}

\usepackage{times}
\usepackage{epsfig}
\usepackage{graphicx}
\usepackage{amsmath}
\usepackage{amssymb}
\usepackage{booktabs}

\usepackage{caption}
\usepackage{xspace}
\usepackage{xcolor}
\usepackage{listings}
\usepackage{enumitem}

\newcommand{\qheading}[1]{\noindent\textbf{#1}}

\newcommand{\ourTitle}[0]{Spatio-Temporal Action Detection Under Large Motion}

\newcommand{\supmat}{\mbox{\textbf{Sup.~Mat.}}\xspace}

\newcommand{\slowfast}[0]{\mbox{SlowFast}\xspace}
\newcommand{\pyslowfast}[0]{py\slowfast}

\newcommand{\detectron}[0]{\mbox{Detectron2}\xspace}

\newcommand{\sota}[0]{\mbox{SOTA}\xspace}
\newcommand{\sotalong}[0]{state-of-the-art\xspace}

\newcommand{\tcn}[0]{\mbox{TCN}\xspace}
\newcommand{\maxpool}[0]{\mbox{MaxPool}\xspace}

\newcommand{\fps}[0]{\mbox{FPS}\xspace}
\newcommand{\iou}[0]{\mbox{IoU}\xspace}
\newcommand{\ioulong}[0]{Intersection-over-Union}
\newcommand{\ap}[0]{\mbox{AP}\xspace}
\newcommand{\aplong}[0]{Average Precision\xspace}
\newcommand{\map}[0]{\mbox{mAP}\xspace}
\newcommand{\maplong}[0]{mean Average Precision\xspace}
\newcommand{\fmap}[1]{\mbox{f-mAP#1}\xspace}
\newcommand{\vmap}[1]{\mbox{v-mAP#1}\xspace}

\newcommand{\motionap}[1]{\mbox{Motion\ap#1}\xspace}
\newcommand{\motionmap}[0]{\mbox{Motion-\map}\xspace}

\newcommand{\vmotionmap}[0]{\mbox{Video Motion-\map}\xspace}

\newcommand{\yolovfive}[0]{\mbox{YOLOv5}\xspace}
\newcommand{\deepsort}[0]{\mbox{DeepSort}\xspace}
\newcommand{\yolodeepsort}[0]{\yolovfive-\deepsort\xspace}
\newcommand{\tracker}[0]{\yolodeepsort}

\newcommand{\roialign}[0]{\mbox{RoI-Align}\xspace}

\newcommand{\videoswin}[0]{\mbox{VideoSwin}\xspace}
\newcommand{\mvit}[0]{\mbox{MViT}\xspace}

\newcommand{\resnet}[0]{\mbox{ResNet-50}\xspace}

\newcommand{\aspp}[0]{\mbox{ASPP}\xspace}
\newcommand{\fpn}[0]{\mbox{FPN}\xspace}

\newcommand{\ava}[0]{\mbox{AVA}\xspace}
\newcommand{\multisports}[0]{\mbox{MultiSports}\xspace}
\newcommand{\multisportsshort}[0]{\mbox{MS}\xspace}
\newcommand{\ucftwofour}[0]{\mbox{UCF24}\xspace}

\newcommand{\gflops}[0]{\mbox{GFLOPS}\xspace}

\newcommand{\convnext}[0]{\mbox{ConvNeXt}\xspace}

\newcommand{\mscoco}[0]{\mbox{MS COCO}\xspace}
\newcommand{\Baseline}[0]{\mbox{Baseline}\xspace}

\newcommand{\tfalong}[0]{Temporal Feature Aggregation\xspace}
\newcommand{\motionwise}[0]{Motion-wise\xspace}
\newcommand{\tfa}[0]{\mbox{TFA}\xspace}

\newcommand{\toialignlong}[0]{Track-of-Interest Align}
\newcommand{\toialign}[0]{\mbox{TOI-Align}\xspace}

\newcommand{\taadlong}[0]{Track Aware Action Detector\xspace}
\newcommand{\taad}[0]{\mbox{TAAD}\xspace}
\newcommand{\TAAD}[0]{\taad}

\newcommand{\lmotion}[0]{\mbox{Large-motion}\xspace}

\newcommand{\myarraystretch}[0]{1.10}

\newcommand{\spatiotemporal}[0]{spatio-temporal\xspace}
\newcommand{\Spatiotemporal}[0]{Spatio-temporal\xspace}
\newcommand{\colorRef}[1]{\textcolor{black}{#1}} 
\newcommand{\reffig}[1]{\colorRef{Fig.~\ref{#1}}}
\newcommand{\refFig}[1]{\colorRef{Figure~\ref{#1}}}

\newcommand{\refsec}[1]{\colorRef{Sec.~\ref{#1}}}

\usepackage{pifont}
\newcommand{\cmark}{\color{green}\ding{51}}
\newcommand{\xmark}{\color{red}\ding{55}}

\def\wacvPaperID{614} 

\wacvfinalcopy 

\ifwacvfinal
\usepackage[breaklinks=true,bookmarks=false]{hyperref}
\else
\usepackage[pagebackref=true,breaklinks=true,colorlinks,bookmarks=false]{hyperref}
\fi

\usepackage[capitalize]{cleveref}
\crefname{figure}{\colorRef{Fig.}}{\colorRef{Figs.}}
\Crefname{figure}{\colorRef{Figure}}{\colorRef{Figures}}
\crefname{section}{\colorRef{Sec.}}{\colorRef{Secs.}}
\Crefname{section}{\colorRef{Section}}{\colorRef{Sections}}
\Crefname{table}{\colorRef{Table}}{\colorRef{Tables}}
\crefname{table}{\colorRef{Tab.}}{\colorRef{Tabs.}}

\usepackage[numbers,sort]{natbib}

\begin{document}

\title{\ourTitle}

\author{Gurkirt Singh
\and
Vasileios Choutas
\and
Suman Saha
\and
Fisher Yu
\and
Luc Van Gool\\
Computer Vision Lab, ETH Zürich
}

\maketitle
\thispagestyle{empty}

\begin{abstract}

Current methods for \spatiotemporal action tube detection often extend a bounding box proposal at a given key-frame into a 3D temporal cuboid and pool features from nearby frames. However, such pooling fails to accumulate meaningful
\spatiotemporal 
features if the position or shape of the actor shows large 2D motion and variability through the frames, due to large camera motion, large actor shape deformation, fast actor action and so on. In this work, we aim to study the performance of cuboid-aware feature aggregation in action detection under large action.
Further, we propose to enhance actor feature representation under large motion by tracking actors and performing temporal feature aggregation along the respective tracks. 
We define the actor motion with intersection-over-union (\iou) between the boxes of action tubes/tracks at various fixed time scales. 
The action having a large motion would result in lower \iou over time, and slower actions would maintain higher \iou. 
We find that track-aware feature aggregation consistently achieves a large improvement in action detection performance, especially for actions under large motion compared to cuboid-aware baseline. 
As a result, we also report state-of-the-art on the large-scale \multisports dataset. The Code is available at \url{https://github.com/gurkirt/ActionTrackDetectron}.

\end{abstract}


\section{Introduction}
\label{sec:intro}

\begin{figure}[t] 
    \centering
    \includegraphics[width=0.48\textwidth]{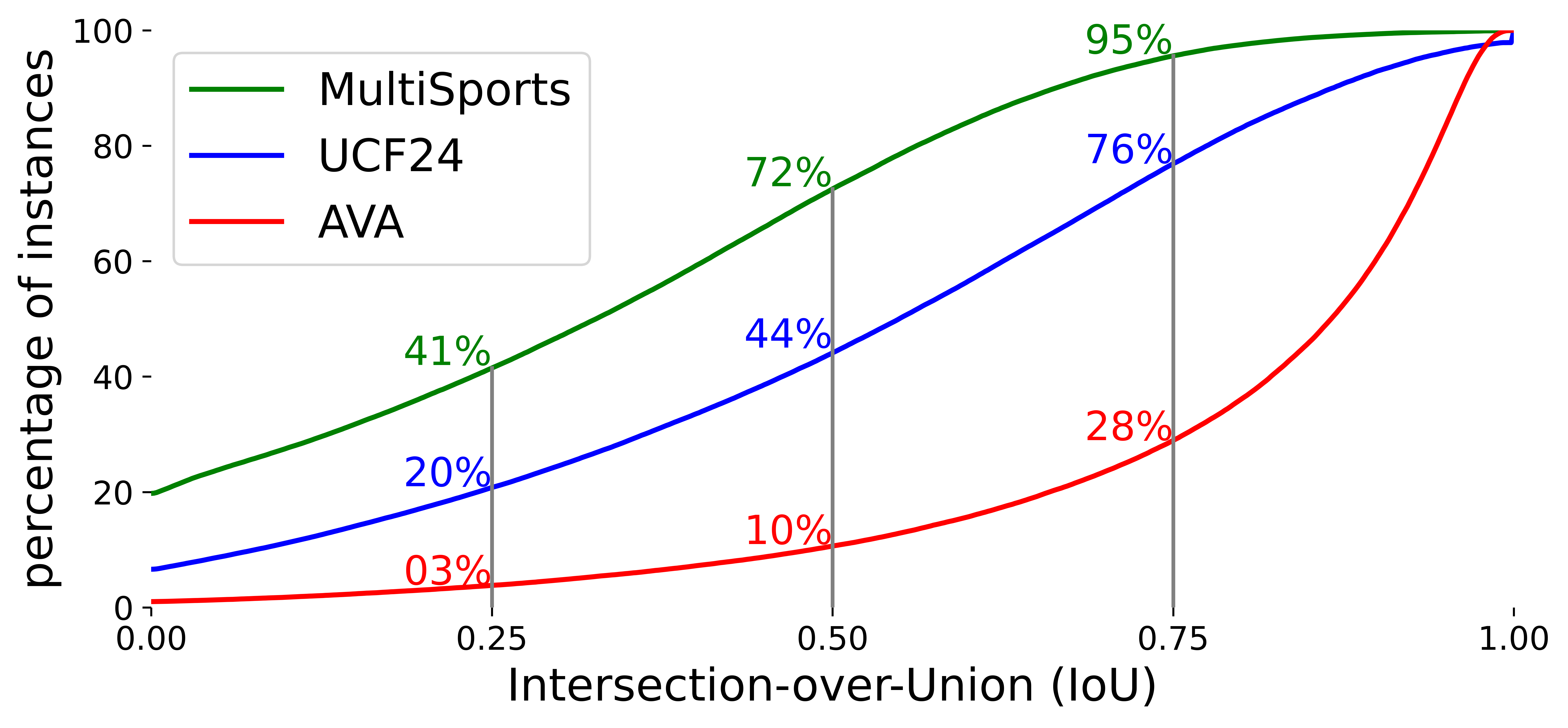}
    \caption{Cumulative density function of \iou measurements for ground-truth bounding box  pairs taken one second apart in the training sets of \ava, \ucftwofour, and \multisports, plotted as percentage of instances falling in cumulative bins shown on the Y-axis. For example, $20\%$ of \multisports instances has an \iou less than or equal to $0.0$ signifying that $20\%$ of instances has very large motion present.
    In contrast, only $10\%$ of \ava instances has an \iou less than $0.5$, meaning that $90\%$ of its instances have a large overlap after one second, \ie large amount instance has small motion. 
    }
    \label{fig:dataset_movement_stats_cumulative} 
    \vspace{-3mm}
\end{figure}

\Spatiotemporal action detection, which classifies and localises actions in space and time,
is gaining attention, thanks to the AVA~\cite{gu2018ava} and UCF24~\cite{soomro2012ucf101} datasets.
However, most of the current state-of-the-art works~\cite{li2020actionsas,singh2017online,feichtenhofer2019slowfast,pan2021actor,tuberZhao2022} focus on pushing action detection performance 
usually by complex context modelling~\cite{tuberZhao2022,pan2021actor,tang2020asynchronous}, larger backbone networks~\cite{feichtenhofer2020x3d,li2022mvitv2,liu2022video}, or by incorporating an optical flow~\cite{zhao2019dance,singh2017online} stream. 
The above methods use cuboid-aware temporal pooling for 
feature aggregation.
In this work, we aim to study cuboid-aware action detection under varying degrees of action instance motion using the \multisports~\cite{li2021multisports} dataset which contains instances with large motions, unlike AVA~\cite{gu2018ava} as shown in \cref{fig:dataset_movement_stats_cumulative}. 

\begin{figure*}[ht] 
    \centering
    \includegraphics[width=1.0\textwidth]{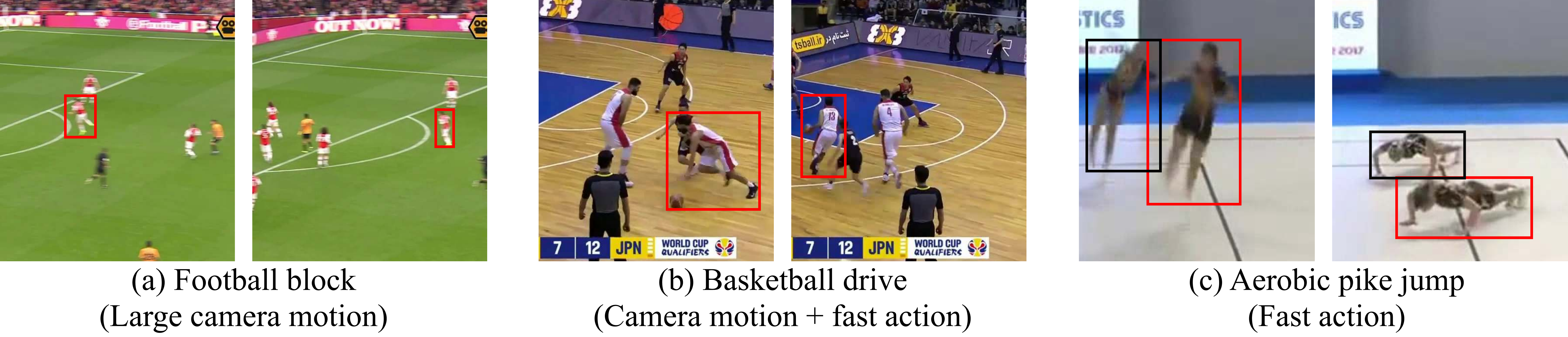}
    \caption{Reasons for large motions: (a) large camera motion (b) camera motion plus actor motion (c) static camera but super fast action. Note that, (b) shows camera zoom out and translation at the same time, and (c) shows Pike-jump action which involves jumping from standing position to air while bringing head and knee close to each other, then lending in horizontal shape on the ground, all this in close to one second. All these images contain pairs of boxes of the same actor, separated by one second in a tube with $0.0$ IoU.
    }
    \label{fig:reasons} 
\end{figure*}

\begin{figure*}[ht] 
  \centering
  \includegraphics[width=1.0\textwidth]{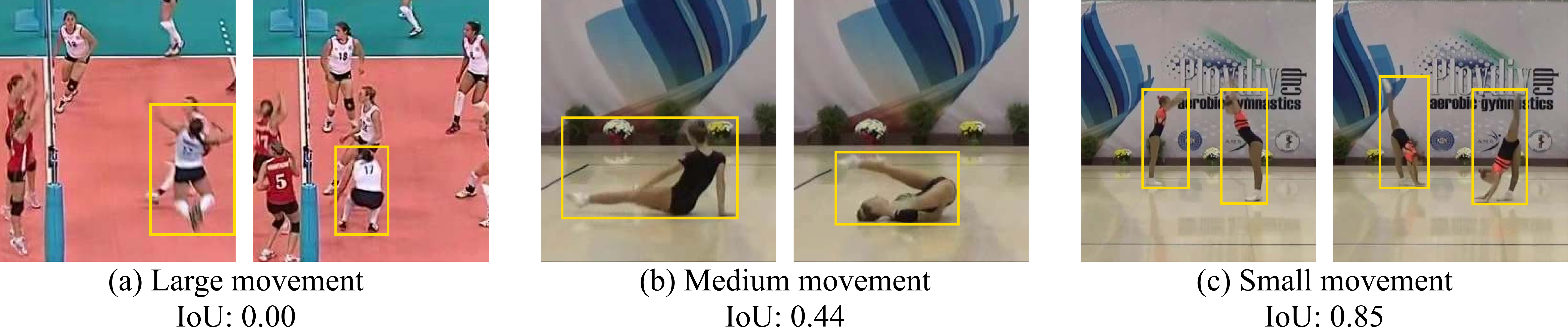}
  \caption{Varying degrees of motion observed for actors in bounding boxes with one second time window with mostly static camera: (a) large motion where actor performs spiking action, results in IoU of $0.0$, meaning large-motion. (b) some pose change as resulting in $0.44$ IoU, meaning medium-motion. (c) change in body pose at same location with IoU being close to $0.85$, i.e. small-motion.} 
  \label{fig:movement_types} 
\end{figure*}

Large object motion can occur for various reasons, e.g., fast camera motion, fast action, body shape deformation due to pose change, or mixed camera and action motions. These reasons are depicted in \cref{fig:reasons}. 
Furthermore, the speed of motions within an action class can vary because of a mixture of the above reasons and the nature of the action type, e.g., pose based or interaction based action. 
Either of these reasons can cause sub-optimal feature aggregation and lead to errors in action classification of a given reason.

We propose to split actions into three categories: Large-motion, medium-motion, and small-motion,
as shown in \cref{fig:dataset_movement_stats_cumulative,fig:movement_types}.
The distinction is based on the \iou of boxes of the same actor over time,
which we can compute using the ground truth tubes of the actors.
We propose to study the performance on different motion categories 
of a baseline cuboid-aware method, without further bells and whistles like context features~\cite{pan2021actor,ning2021person,tang2020asynchronous} or long-term features
\cite{wu2019long,tang2020asynchronous}, because large-motion happens quickly in a small time window,
as seen in \cref{fig:dataset_movement_stats_cumulative} and \ref{fig:reasons}.
In large-motion cases the \iou would be small (\cref{fig:movement_types} (a)), and as a result
a 3D cuboid-aware feature extractor will not be able to capture
features centred on the actor's location throughout the action.
To handle the large-motion case, we propose to track the actor over time and extract features using \toialignlong{ }(\toialign); resulting in \taadlong (\taad).
Further, we study different types of feature aggregation modules on top of TOI-Aligned features for our proposed \taad network, shown in \cref{fig:main_figure}. 

To this end, we make the following contributions:
(a) we are the first to study large-motion action detection systematically, using evaluation metrics for each type of motion, similar to object detection studies on \mscoco~\cite{lin2014microsoft} based on object sizes. 
(b) we propose to use tube/track-aware feature aggregation modules to handle large motions, and we show that this type of module helps in achieving great improvements over the baseline, especially for instances with such large motion. 
(c) in the process, we set a new state-of-the-art for the \multisports dataset by beating last year's challenge winner by a substantial margin.

\section{Related Work}
\label{sec:related_work}

Action recognition~\cite{carreira2017quo,wang2018nonlocal,feichtenhofer2019slowfast,weinzaepfel2021mimetics,feichtenhofer2020x3d,singh2019recurrent,liu2022video,li2022mvitv2} models provide strong video representation models. However, action recognition as a problem is not as rich as action detection, where local motion in the video needs to be understood more precisely. Thus, action detection is the more relevant problem for understanding actions under large motion.

We are particularly interested in the \spatiotemporal~\emph{action detection}
problem \cite{Georgia-2015a,gu2018ava,girdhar2018better,wu2019long,tuberZhao2022}, 
where an action instance is defined as a set of linked bounding boxes over time, called action tube.
Recent advancements in online action detection~\cite{soomro2016predicting,singh2017online,behl2017incremental,kalogeiton2017action,li2020actionsas,yang2019step}
lead to performance levels very competitive with (generally more accurate) offline action detection methods~\cite{gu2018ava,wang2018nonlocal,saha2016deep,saha2017amtnet,peng2016eccv,zhao2019dance,singh2018tramnet,singh2018predicting,van2015apt} on the \ucftwofour~\cite{soomro2012ucf101} dataset. 

\ucftwofour has been a major benchmark for \spatiotemporal action detection (i.e. action tube detection), rather than \ava ~\cite{gu2018ava}. The former is well suited for action tube detection research, as it provides dense action tube annotations, where every frame of the untrimmed videos is annotated (unlike \ava~\cite{gu2018ava}, in which videos are only annotated at one frame per second).
More recently, Li~\etal~\cite{li2021multisports} proposed the \multisports dataset, which resolves two main problems with the \ucftwofour dataset.
Firstly, it has more fine-grained action classes.
Secondly, it has multiple actors performing multiple types of action
in the same video. 
As a result, the \multisports dataset is comparable to \ava in terms of diversity and scale. 
Moreover, the \multisports dataset is densely annotated, every frame at a rate of 25 frames per second, 
which makes it ideal to understand action under large motion, as shown in Fig.~\ref{fig:dataset_movement_stats_cumulative}.

At the same time, there have been many interesting papers~\cite{feichtenhofer2019slowfast,feichtenhofer2020x3d,tang2020asynchronous,pan2021actor,chen2021watch} that focus on keyframe based action detection on \ava \cite{gu2018ava}. 
\ava has been helpful in pushing action detection research on three fronts. Firstly, backbone model representations are much better now thanks to works like~\cite{feichtenhofer2019slowfast,feichtenhofer2020x3d,liu2022video,wang2018nonlocal,chen2021watch}. 
Secondly, long-term feature banks (LBF) ~\cite{wu2019long} came to the fore~\cite{zhang2019structured,tang2020asynchronous,pan2021actor}, capturing some temporal context, but without temporal associations between actors. 
Thirdly, interactions between actors and object have been studied~\cite{pan2021actor,zhang2019structured,tang2020asynchronous,ning2021person}. 
Once again, the problem we want to study is action detection under large motion, which happens quickly at a small temporal scale. All the above methods use cuboid-aware pooling for local feature aggregation, which - as we will show - is not ideal when the motion is quick and large.
As a result, we borrow the \slowfast~\cite{feichtenhofer2019slowfast} network
as the baseline network for its simplicity and \spatiotemporal representational power.
Also, it has been used for \multisports~\cite{li2021multisports} as baseline
and in many other works on \ucftwofour as a basic building block.

The work of Weinzaepfel~\etal~\cite{weinzaepfel2015learning}
is the first to use tracking for action detection.
That said, their goal was different than ours.
They used a tracker to solve the linking
problem in the tube generation part,
where action classification was done on a frame-by-frame basis given the bounding box proposals from tracks. We, on the other hand,
propose action detection by pooling features from within
entire tracks. 
Gabriellav2~\cite{dave2022gabriellav2} is another method that makes use of tracking to solve the problem of temporal detection of co-occurring activities, but it relies on background subtraction which would fail in challenging in-the wild videos.
Singh~\etal~\cite{singh2018tramnet},
Li~\etal~\cite{li2020actionsas} and Zhao~\etal~\cite{tuberZhao2022} are the only works generating flexible micro-tube proposals without the help of tracking. However, these approaches are limited to a few frames (2-10).
Without the possibility to scale to larger time windows of 1-2 seconds as required for multi-frame tube anchors/query to regress box coordinates on a large number of frames, performance drops after a few frames. 


\section{Methodology}
\label{sec:method}
\begin{figure*}[t] 
  \centering
  \includegraphics[width=0.95\textwidth]{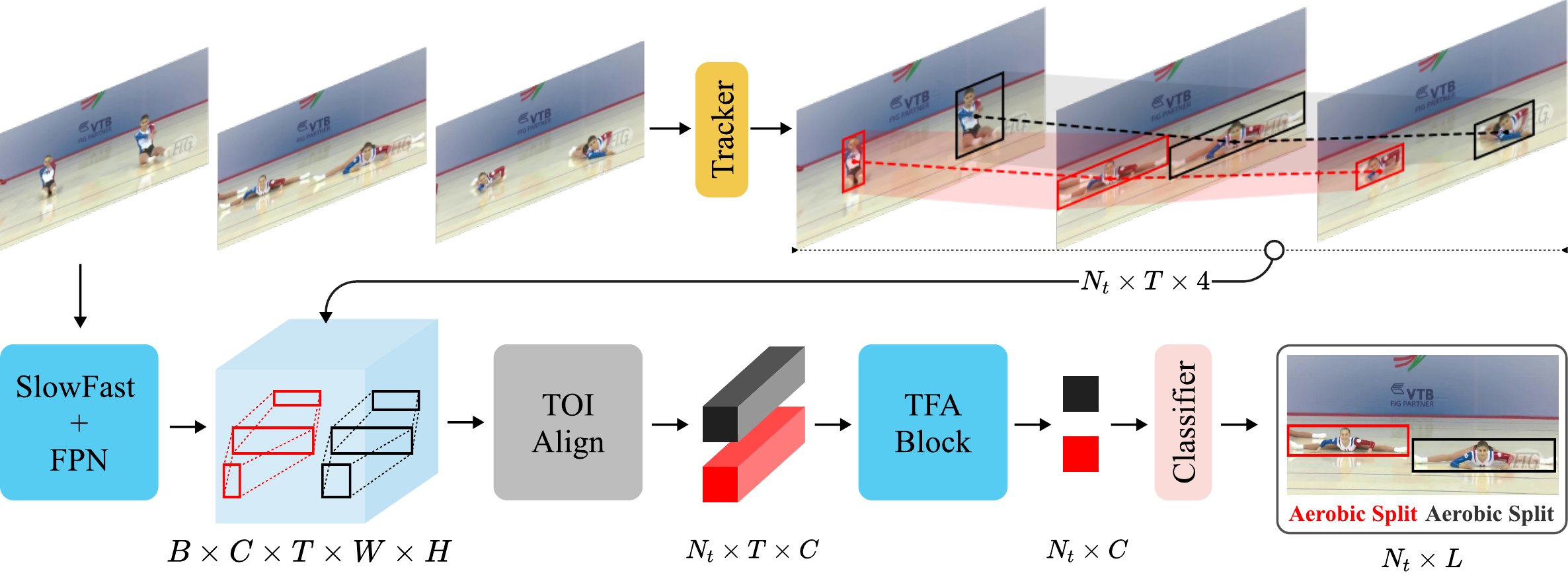}
  \caption{%
  Proposed \taadlong~(\taad):
  Given an input clip with $T$ frames, we extract features
  using a video recognition network \cite{feichtenhofer2019slowfast}
  and $N_t$ per-actor tracks from a tracker.
  The \toialign operation extracts per-track features
  from the entire video sequence, using an \roialign operation and the
  track boxes, returning a $N_t \times T \times C$ feature array.
  Next, the \tfalong (\tfa) module aggregates the features along the
  temporal dimension and passes the resulting $N_t \times C$ array
  to the action classifier that predicts the action label.
  }
  \label{fig:main_figure} 
\end{figure*}

In this section, we describe the proposed method to handle actions
with large motions, which we call \taadlong (\taad).
We start by tracking actors in the video,
using a tracker described in Section~\ref{subsec:tracker}.
At the same time, we use a neural network 
designed for video recognition, 
\slowfast \cite{feichtenhofer2019slowfast},
to extract features from each clip.
Using the track boxes and video features, we pool per-frame features
with a \roialign operation \cite{he2017mask}.
Afterwards, a \tfalong (\tfa) module
receives the per-track features and 
computes a single feature vector,
from which a classifier predicts the final action label.
\refFig{fig:main_figure} illustrates
each step of our proposed approach.

\subsection{Baseline Action Detector}
\label{subsec:backbone}

We select a \slowfast \cite{feichtenhofer2019slowfast} network
as our video backbone.
The first reason for this choice is that its performance
is still competitive to larger scale transformer models,
such as \videoswin \cite{liu2022video} or
\mvit \cite{fan2021multiscale,li2022mvitv2}, on the task of
\spatiotemporal action detection.
Furthermore, \slowfast is computationally more efficient
than the transformer alternatives, with a cost of 
65.7 \gflops compared to 88, at least, and 170 
for \videoswin \cite{liu2022video} and \mvit \cite{fan2021multiscale} respectively,
and offers features at two different temporal scales.
Having different temporal scales is important, especially since
we aim at handling fast and/or large motions, where a smaller
scale is necessary.
Finally, \slowfast is the default backbone network of choice for the
\multisports and \ucftwofour datasets, which are the main benchmarks in this work, facilitating comparisons with existing work.

We implement our baseline using \pyslowfast\cite{fan2020pyslowfast} with a 
\resnet~\cite{he2016deep} based \slowfast~\cite{feichtenhofer2019slowfast}
architecture, building upon the works of 
Feichtenhofer \etal \cite{feichtenhofer2019slowfast}
and Li \etal \cite{li2021multisports}. 
First, we add background frames (+bg-frames),
i.e. frames erroneously detected by our detector,
\yolovfive, as extra negative samples for training the action detector.
Next, we replace the
multi-label with a multiclass classifier, switching
from a binary cross entropy per class to a cross entropy loss (CE-loss).
Finally, we also added a downward \fpn block (see Sup.Mat. for details).
Through these changes, we aimed to build the strongest possible baseline.

\subsection{Tracker}
\label{subsec:tracker}

We employ a class agnostic version of  
\tracker~\cite{yolov5deepsort2020} as our tracker,
which is based on \yolovfive \cite{yolov5,redmon2016you} and
TorchReID\cite{torchreid}. 
We fine-tune the medium size version of \yolovfive as the detection model for `person` classes. 
A pretrained OsNet-x0-25~\cite{zhou2021osnet} is used as re-identification (ReID) model. 
As we will show in the experiment section,
a tracker with high recall, \ie small number of missing associations,
is key for improving performance action tube detection. 
We will also show that fine-tuning the detector is a necessary step,
particularly for \ucftwofour, where the quality and resolution of the videos
is small.

The tracker can also be used as bounding box proposal filtering module. 
Sometimes, the detector produces multiple high scoring detections which are spurious and lead to false positives but these detections do not match to any of the tracks being generated because they are not temporally consistent. 
The proposals generated by tracks can be used with the baseline methods at test-time. This helps improve the performance of the baseline method. 


\subsection{Temporal Feature Aggregation}
\label{subsec:tpa}

\qheading{\toialignlong{ }(\toialign):} The \slowfast video backbone processes the input clip and produces
a $T \times H \times W$ feature tensor, while our tracker returns
an array with size $N_t \times T \times 4$ that contains the boxes
around the subjects. An \roialign \cite{he2017mask} takes these two arrays as input
and produces a feature array of size $N_t \times T \times H \times W$, \ie
one feature tube per track. In case the length of the track is smaller than the length of the input clip, we replicate the last available bounding box in the temporal direction, which occurs around $~3\%$ of input clips in \multisports dataset.

\qheading{Feature aggregation:} In order to predict the label of a bounding box in a key-frame,
we need to aggregate features across space and time. First we apply average pooling in spatial dimensions on features extracted by \toialign, then the \tfalong role is performed by 
one of the following variants considered: 

\setlist{itemsep=0pt}
\begin{enumerate}
    \item Max-pooling over the temporal axes (\maxpool).
    \item A sequence of temporal convolutions (\tcn).
    \item A temporal variant of Atrous Spatial Pyramid Pooling (\aspp) \cite{deeplabv3plus2018}. We modify \detectron{'s} \cite{wu2019detectron2} ASPP implementation, replacing 2D with 1D convolutions. 
\end{enumerate}


We also tried a temporal version of \convnext~\cite{liu2022convnet} and \videoswin~\cite{liu2022video} blocks, however these resulted in unstable training,
even with the tunning of learning rates and other hyperparameters. In our experiments, we only used one layer of temporal convolution for our \tcn module, adding more layers did not help. See the \supmat for more details.

\subsection{Tube Construction}
\label{subsec:tube_con}

Video-level tube detection requires the construction of action tubes from per-frame detections.
This process is split into two steps~\cite{saha2016deep}. The first links the proposals to form tube hypotheses (i.e. action tracks). The second trims these hypotheses to the part where there is an action.
One can think of these two steps as a tracking step plus a temporal (start and end time) action detection step.
The majority of the existing action tube detection methods \cite{li2021multisports,singh2022road,li2020actionsas,singh2018tramnet}
use a greedy proposal linking algorithm first proposed by in~\cite{singh2017online,kalogeiton2017action} for the first step. 
For the baseline approach, we use the same method for the tube linking process from~\cite{singh2017online}.
Since for our method (\taad) we already have tracks, the linking step is already complete. 
The temporal trimming of action tracks is performed using label smoothing optimisation~\cite{saha2016deep}, which is used by many previous works~\cite{kalogeiton2017action,li2020actionsas}. In particular, we use class-wise the temporal trimming implementation provided by~\cite{singh2017online}.


\subsection{Datasets}
We evaluate our idea on two densely annotated datasets (\multisports~\cite{li2021multisports} and \ucftwofour \cite{soomro2012ucf101}) with frame and tube level evaluation metrics for actions detection, unlike \ava \cite{gu2018ava}, which is sparsely annotated and mostly used for frame level action detection.

\qheading{\multisports~\cite{li2021multisports}} 
is built using 4 sports categories, collecting $3200$ video clips annotated at 25 \fps,
and annotating 37701 action tube instances with $902$k bounding boxes.
Although it contains $66$ action classes, we follow the official evaluation
protocol~\footnote{\url{https://github.com/MCG-NJU/MultiSports/}} that uses $60$ classes. 
Due to the fine granularity of the action labels the length of each action segment
in a clip is short, with an average 
tube length of $24$ frames, equal to one second,
while the average video length is $750$ frames.
Each video is annotated with multiple instances of multiple action classes,
with well defined temporal boundaries.
\multisports contains action instances with large motions around actors, as shown in \reffig{fig:dataset_movement_stats_cumulative}

\qheading{\ucftwofour \cite{soomro2012ucf101}}
consists of $3207$ videos annotated at $25$ \fps with $24$ action classes from different sports,
$4458$ action tube instances with $560$K bounding boxes.
Videos are untrimmed,
with an average video length of
$170$ frames and average action tube length of $120$ frames. 
The disadvantages of \ucftwofour are (1) the presence
of only one action class per video and (2) the low image quality,
due to compression and the small resolution, namely $320 \times 240$ pixels.
Even though \ucftwofour has less diversity,
less motion, fewer classes and more labelling noise compared to \multisports,
it is still useful to evaluate action detection performance,
thanks to its temporally dense annotations.


\subsection{Implementation Details}
\label{subsec:training_hparams}
We use 32 frames as input with sampling rate of 2, which means more 2 seconds of video clip. 
We use Slowfast-R50-$\tiny{8\times8}$~ \cite{feichtenhofer2019slowfast}, meaning  speed ratio $\alpha=8$ and channel ratio $\beta=1/8$.
We use stochastic gradient descent (SGD) to optimise the weights, with a learning rate of 
$0.05$ and batch size of $32$ on $4$ GPUs. 
We use 1 epoch to warm up the learning rate linearly,
followed by a cosine learning rate schedule \cite{cosine_lr}, with a final learning rate
of $0.0005$, for a total of 5 epochs. 
Note that we only train for $3$ epochs on \ucftwofour.
All our networks are trained with a batch size equal to $32$ on $4$ Titan X GPUs.
We use the frame-level proposal released by \cite{li2021multisports} for \multisports, for the fairness of comparison. More details can be found in \supmat.


\section{Experiments}
\label{experiments}

In this section, we evaluate our \TAAD method along with \tfa modules on the  \multisports and \ucftwofour datasets. We start by defining the metrics used in \refsec{subsec:metrics}
and motion category
classification in \refsec{subsec:motion_classifcation}. 
Firstly, we study the impact of different \tfa modules under different motion conditions in \refsec{subsec:tfa}.
Secondly, we compare our \TAAD method with \sotalong methods in \refsec{subsec:comparisons}. 
Later, we discuss the baseline model and the impact the tracker has in \refsec{subsec:baseline_exp}. 
We finish with a discussion in section~\refsec{subsec:discussion}.

\subsection{Metrics}
\label{subsec:metrics}

We report metrics that measure our detector's performance both
at frame- and video-level, computing frame and video \maplong (\map),
denoted as \fmap{} and \vmap{} respectively.
These metrics are common in action detection works~\cite{kalogeiton2017action,weinzaepfel2015learning,li2020actionsas}.
A detection is correct if and only if its \ioulong (\iou) with a ground-truth
box or tube, for frame and video metrics respectively, is larger than a given threshold (e.g. $0.5$) and
the predicted label matches the ground-truth one. 
From this, we compute the
\aplong (\ap) for each class and the mean across classes, to get the desired \map metric.
Tube overlap is measured by \spatiotemporal-\iou proposed by ~\cite{weinzaepfel2015learning}, 
similar to ~\cite{li2021multisports}, we use the ACT\footnote{\url{https://github.com/vkalogeiton/caffe/tree/act-detector}} evaluation code.


\subsection{Motion categories} 
\label{subsec:motion_classifcation}
We split actions into three motion categories: large, medium and small.
Computing per-motion-category metrics requires labelling the ground-truth
action tubes. We start this process by computing the \iou
between a pair of boxes separated by offsets equal to $[4,8,16,24,36]$ in sliding window fashion. 
We average these 5 \iou values and get the final \iou value as a measure of speed.
We then split the dataset into three bins of equal size.
We can then assign a 'large, medium, or small' motion label to each instance:
\begin{equation}
    \text{\multisports} = \begin{cases}
        \text{Large} , & \quad \text{\iou} \in [0.00, 0.21] \\
        \text{Medium} , & \quad \text{\iou} \in [0.21, 0.51] \\
        \text{Small} , & \quad \text{\iou} \in [0.51, 1.00]
      \end{cases} \\
\end{equation}
\begin{equation}
  \text{\ucftwofour} = \begin{cases}
        \text{Large} , & \quad \text{\iou} \in [0.00, 0.49] \\
        \text{Medium} , & \quad \text{\iou} \in [0.49, 0.66] \\
        \text{Small} , & \quad \text{\iou} \in [0.66, 1.00]
      \end{cases} 
\end{equation}
Given these labels, 
we can compute \ap metrics per motion category.
There are two options for these metrics. 
The first is to compute the
\ap for large, medium and small motions per action class and then
average across actions. 
We call this metric \emph{\motionmap}.
The alternative is to ignore action classes and compute the \ap
for large, medium and small motions, irrespective of the action,
which we call \emph{\motionap}. This essentially measures action
detection accuracy \wrt to motion speed, irrespective of class.
We compute the metrics both on a per-frame and on a per-video level, following the two methods just described. Video metrics are denoted with a \emph{video} prefix.
We will release the code for training and testing our \taad network along with evaluation scripts for both \motionap{} and \motionmap.

\begin{table*}[t!]
    \centering
    \scriptsize
    \renewcommand{\arraystretch}{\myarraystretch}
    \setlength{\tabcolsep}{6pt}
    
    \caption{
    \motionwise ablation of \tfalong modules. We investigate the effect of different feature aggregation modules
    using frame- and video-\map to measure model performance,
    both with the classic definition and with our proposed motion categories.
    Aggregating features across tracks, instead of cuboids,
    improves action detection performance across all categories, with a particularly noticeable
    improvement for large motions.
    For example, the \tcn module improves large motion \motionmap by $8.4$,
    with an improvement of only $4.5$ points for small motions.
    }
    \resizebox{0.96\linewidth}{!}{
    \begin{tabular}{lccccc ccc}
    \toprule
     & \fmap{@0.5} & \multicolumn{3}{c}{\motionmap{@0.5}} &
    \vmap{@0.5} & \multicolumn{3}{c}{\vmotionmap{@0.5}} \\
    Method &  & Large & Medium & Small &  & Large & Medium & Small \\
    \midrule
     & \multicolumn{8}{c}{\multisports \cite{li2021multisports} } \\ 
    \midrule
    Baseline (SlowFastR50~\cite{feichtenhofer2019slowfast}) & 49.6 & 36.5 & 49.5 & 54.9 & 31.2 & 14.2 &     33.6 &     45.1  \\
    Baseline + track${}^{\dagger}$ & 50.6 & 39.7 & 50.1 & 56.3 & 33.0 & 15.4 &     34.7 &     45.7  \\ 
    \taad + \maxpool & 53.9  & 43.8 & 52.7 & 57.7 &  34.8 & 16.7 &     35.5 &     47.4 \\
    \taad + \aspp & 54.4  & 44.2 & 52.9 & 58.4 &  36.0 & \textbf{18.8} &     37.5 &     46.0 \\
    \taad + \tcn & \textbf{55.3} & \textbf{44.9} & \textbf{53.4} & \textbf{60.4} &  \textbf{37.0} & 17.9 &     \textbf{38.1} &     \textbf{47.3} \\
    \midrule
     & \multicolumn{8}{c}{\ucftwofour \cite{soomro2012ucf101} } \\ 
     \midrule
    Baseline (SlowFastR50~\cite{feichtenhofer2019slowfast}) & 75.9 & 67.0 & 77.3 & 70.6 & 45.4 & 33.3 &  47.0 &  46.0  \\
    Baseline + track${}^{\dagger}$ & 78.3 & 68.6 & 79.0 & 72.1 & 47.4 & 34.8 &  47.9 &  50.7  \\ 
    \taad + \tcn & \textbf{81.5} & \textbf{74.9} &
    \textbf{83.7} & \textbf{75.1}  &  \textbf{52.0}  &  \textbf{38.3}  & \textbf{51.2} &   \textbf{50.2} \\
    \bottomrule
    \multicolumn{9}{l}{${}^{\dagger}$tracks used a filtering module at frame-level and tube construction module at video-level.}
    \end{tabular}
    }
    \label{tab:splits} 
\end{table*}

\subsection{Motion-wise (main) results}
\label{subsec:tfa}

As the main objective for this work, we first study how the cuboid-aware baseline compares against our track-based \taad under significant motion. 
We compare different choices for
temporal feature aggregation. In \cref{tab:splits}, we measure the frame- and video-motion-\map,
for models trained with
different \tfa{s}, on \multisports and \ucftwofour.
Pooling features across tracks, instead of neighbouring frames,
even with a relatively simple pooling strategy, \ie Max-Pool over the \spatiotemporal dimensions,
results in stronger action detectors, with a 5.7 \%
 and 
5.8 \%
frame
and video
\map
boost on \multisports.
More involved feature aggregation
strategies, such as the temporal convolution blocks (\tcn) or \aspp variant, lead to further
gains. Note that the biggest improvements on \multisports occur in the large motion category,
+ 8.4 \%  \motionmap{}, with smaller gains in medium (+3.9\%)
and small (+5.5 \%) motions.

\begin{table}[t!]
    \centering
    \footnotesize
    \renewcommand{\arraystretch}{\myarraystretch}
    \setlength{\tabcolsep}{8pt}
    
    \caption{
    Motion-wise ablation with \motionap{} metric. We investigate the effect of different \tfa modules
    using frame-level \motionap{} to asses the quality of motion-wise action detection in comparison to baseline on \multisports dataset.
    }
    \begin{tabular}{llll}
    \toprule
     & 
     \multicolumn{3}{c}{\motionap{@0.5}} \\
    Method 
    &  Large & Medium & Small \\
    \midrule
    Baseline 
    & 63.2 & 77.7 & 82.4   \\
    Baseline + track${}^{\dagger}$ 
    & 64.6\small{(+1.5)} & 78.7\small{(+1.0)} & 84.4\small{(+2.0)}  \\ 
    \taad+\maxpool 
    & 70.2\small{(+7.0)} & 83.4\small{(\textbf{+5.7})} & 86.1\small{(+3.9)}  \\
    \taad+\aspp 
    & 71.1\small{(\textbf{+7.9})}  & 83.4\small{(\textbf{+5.7})} & 86.9\small{(+4.5)}  \\
    \taad+\tcn 
    & 70.4\small{(+7.2)} & 83.3\small{(+5.6)} & 87.3\small{(\textbf{+4.9})}\\
    \bottomrule
    \multicolumn{4}{l}{${}^{\dagger}$ tracks used as filtering module.
    }
    \end{tabular}
    \label{tab:motion_ap} 
\end{table}
\cref{tab:motion_ap} contains \motionap{} results on \multisports
for different \tfa module choices.
It is clear that \taad combined with any of the \tfa modules leads to
large performance gains. 
Larger motions benefit the most, followed by medium
and small motions.
For example, the \aspp module helps more with large motions
($+7.9$) than with small motions ($+4.5$).
We observe the same trend
in \cref{tab:splits},
both for frame and video \motionmap. 

These results signify that there is a large gap between the performance for large vs. small motion action instances for the baseline method. The combination of
\taad with any of the \tfa modules helps to reduce this discrepancy and improves the overall performance for both datasets.

\begin{table}[t]
    \centering
    
    \footnotesize
    \renewcommand{\arraystretch}{\myarraystretch}
    \setlength{\tabcolsep}{8pt}
    
    \caption{%
    Comparison of action detection performance 
    of the proposed method to our baseline model and other
    state-of-the-art methods on \multisports dataset.
    \taad combined with \tfa modules leads to \sotalong detection performance.
    }
    \begin{tabular}{lcccc}
    \toprule
    Method &  \fmap{} & \multicolumn{3}{c}{\vmap{}} \\
    &  0.5 & 0.2  & 0.5 & .1:.9 \\\midrule
    YOWO \cite{li2020actionsas,li2021multisports}  & 25.2  & 12.9 & 9.7 & -- \\
    MOC \cite{li2020actionsas,li2021multisports}  & 25.2  & 12.9 & 9.7 & -- \\
    \slowfast-R50 \cite{feichtenhofer2019slowfast,li2021multisports}  & 27.7  & 24.2 & 9.7 & --   \\
    \slowfast-R101 \cite{ning2021person}  & 29.5  & 28.1 & 8.4 & 12.3   \\
    \slowfast-R101+PCCA \cite{ning2021person}  & 42.2  & 41.0 & 20.0 & 20.9  \\
    \Baseline (ours)  & 49.6 & 54.1 & 31.3 & 28.9 \\
    \Baseline + tracks (ours) ${}^{\dagger}$   & 50.6 & 56.3 & 33.0 & 30.9 \\
    \taad + \maxpool (ours)  & 53.9 & 58.6 & 34.8 &  32.4 \\
    \taad + \aspp (ours)  & 54.4 & 59.2 & 36.0 &  33.0 \\
    \taad + \tcn (ours)  & \textbf{55.3} & \textbf{60.6} & \textbf{37.0} &  \textbf{33.7} \\
    \bottomrule
    \multicolumn{5}{l}{ ${}^{*}$ evaluated using tracks at test time.}
    \end{tabular}
    \label{tab:multisports} 
\end{table}
\begin{table}[t]
    \centering
    \footnotesize
    \renewcommand{\arraystretch}{\myarraystretch}
    \setlength{\tabcolsep}{6pt}
    
    \caption{Comparison of action detection performance (\fmap{} and \vmap{}) of the proposed method along with our baseline model and other \sota methods on UCF24 dataset. \taad with \tcn shows performance gain compared to baseline, with competitive performance to other methods that are specifically designed for \ucftwofour,
    including spatial context~\cite{tuberZhao2022,pan2021actor} module and sophisticated transformer head used by~\cite{tuberZhao2022}.
    \taad is even better than some of the approach that us optical flow (``F'') stream as input along with visual stream (``V'').
    }
    \begin{tabular}{lccccc}
    \toprule
    Methods  & Input & \fmap{} & \multicolumn{3}{c}{\vmap{}} \\
    &  &  & 0.2  & 0.5 
    & 0.5:0.9 \\
    \midrule
    ROAD \cite{singh2017online}    & V+F & --      & 76.4 & 45.2  
    & 20.1 \\
    AMTnet \cite{saha2017amtnet}    & V+F     & -- & 78.5  & 49.7 
    & 24.0 \\
    ACT \cite{kalogeiton2017action} & V+F & 67.9  & 76.5 & 49.2  
    & 23.4  \\
    TACNet \cite{song2019tacnet}  & V+F & 72.1  & 77.5 & 52.9 
    & 24.1  \\
    FlowDance \cite{zhao2019dance}  & V+F & --  & 78.5 & 50.3 
    & 24.5  \\
    I3D \cite{gu2018ava} & V+F & 76.3  & -- & 59.9 
    & -- \\
    MOC \cite{li2020actionsas} & V+F  & 78.0  & 82.8 & 53.8 
    & 28.3  \\
    TubeR \cite{tuberZhao2022} & V+F  & 81.3  & \textbf{85.3} & \textbf{60.2} 
    & \textbf{29.7}  \\
    \midrule
    YOWO \cite{kopuklu2019you} & V  & 78.0  & 75.8 & 48.8 
    & -- \\
    TubeR \cite{tuberZhao2022} ${}^{*}$ & V  & 80.1  & \textbf{82.8} & \textbf{57.7} 
    & \textbf{28.7}  \\
    \Baseline & V  & 75.8  & 76.7 & 45.5 
    & 19.7 \\
    \Baseline + tracks ${}^{\dagger}$ & V  & 78.8  & 77.4 & 47.4 
    & 20.2  \\
    \taad+\tcn & V  & \textbf{81.5}  & 79.6 & 52.0 
    & 23.0  \\
    \bottomrule
    \multicolumn{6}{l}{ ${}^{\dagger}$ evaluated using tracks at test time.}\\
    \multicolumn{6}{l}{ ${}^{*}$ TubeR uses large transformer head plus complex context modelling.}
    \end{tabular}
    \vspace{-3mm}
    \label{tab:ucf24} 
\end{table}
\subsection{Comparison to the State-of-the-art}
\label{subsec:comparisons}

We compare our proposed detector with the state-of-the-art for \multisports and \ucftwofour, for both frame and tube level action detection, unlike approaches~\cite{pan2021actor,tang2020asynchronous} which solely focus on frame level evaluation.
It is important to note that, similar to the baseline, \taad does not make use of any spatial context. Hence, gains are made using track aware feature aggregation rather than by using other \spatiotemporal context modelling modules~\cite{tuberZhao2022}. 

We report frame and video \map 
for different methods in \cref{tab:multisports}, namely \slowfast variants from the original \multisports paper,
Ning \etal's \cite{ning2021person}
Person-Context Cross-Attention Modelling network and our improved baseline,
and three versions of our model, the
one with \maxpool along the temporal dimensions,
the \aspp variant and the temporal convolutional network (\tcn).
\cref{tab:multisports} contains  the results of these experiments, where
we clearly see the benefit of using tracks for action detection.
The addition of feature pooling along tracks,
even with the simpler \maxpool version, outperforms
our improved baseline by 4.3 \% frame \map. Better temporal fusion strategies,
\ie \aspp and \tcn, lead to further benefits. 
As a result, we set a new \sotalong for the \multisports dataset.
Note that all our TFA modules add less than 1M FLOPS ($<2\%$) to the computation time of the whole network.


Finally, we compare our proposed \taad model on the older \ucftwofour dataset in \cref{tab:ucf24}.
Our model outperforms most existing methods, with the exception of TubeR~\cite{tuberZhao2022} and MOC~\cite{li2020actionsas}. 
We think the reason is that TubeR uses a set prediction framework~\cite{carion2020end} with a transformer head (plus 3 layers for each encoder- and decoder-transformer) on top a CNN backbone (CSN-152). Moreover, they use actor context modelling similar to~\cite{pan2021actor}. 
It is also important that I3D based TubeR needs $132M$ FLOPS, which is much higher than the $97M$ needed by SlowFastR5-TCN based \taad. 
MOC uses flow stream as additional input and uses DLA-34\cite{yu2018deep} as backbone network. 
Note that our goal is to analyse and improve action detection
performance across different actor motion speeds. Hence, we do not
use any spatial attention or context modelling~\cite{tuberZhao2022} between actors.
These are certainly very interesting topics, orthogonal to our proposed approach.
This said, our network consistently shows improvements in all metrics for both datasets when compared to our baseline.

Additionally, the low quality on \ucftwofour given by \yolovfive also hampers performance. 
We report the corresponding \yolovfive+DeepSort metrics in 
\cref{tab:trackers_recall}. 
Fine-tuning the detector on each dataset 
is a necessary step,
especially on \ucftwofour where the video quality is worse than \multisports.


\subsection{Building a Strong Baseline on \multisports}
\label{subsec:baseline_exp}

\begin{table}[t]
  \centering
  \footnotesize
  \setlength{\tabcolsep}{6pt}
  \renewcommand{\arraystretch}{\myarraystretch}
  \caption{%
  Baseline progression on \multisports dataset with proposals released by~\cite{li2021multisports}. 
  Adding more negatives proposals from non-action frames, in the form of proposals erroneously detected by the 
  per-frame detector, converting the problem from multi-label to multiclass classification
  and adding a \fpn leads to a more effective action detector.
  }
  \label{tab:baseline_progression}
  \resizebox{1.0\linewidth}{!}{
  \begin{tabular}{lccccc}
      \toprule
      Method & SlowFast\cite{li2021multisports} & SlowFast  & +bgFrames & +CE-loss & +FPN \\ 
      \midrule
      \#keyframes & unknown & 288K & 354K & 354K & 354K \\
      \fmap{@0.5} & 27.7 & 34.5 & 39.7 & 49.0 & 49.6 \\
      \bottomrule
  \end{tabular}
  }
\end{table}
Here, we investigate the effect of our proposed changes on the performance
of the baseline action detector. \cref{tab:baseline_progression} contains the \fmap{@0.5} values,
computed on \multisports,
for our re-implementation of the \resnet \slowfast network, the addition of the background
negative frames, the conversion of the multi-label to a multiclass classification and finally
the addition of the \fpn. Each component improves the performance of the detector, leading to a much
stronger baseline. 

\begin{table}[t]
  \centering
  \captionsetup{font=footnotesize}
  \footnotesize
  \setlength{\tabcolsep}{6pt}
  
  \caption{%
  Recall of 
  Class agnostic \yolovfive based \deepsort tracker on
  \multisportsshort and \ucftwofour dataset,
  with and without fine-tuning the detector on each dataset. 
  Even though \multisports is more complex, the tracker has better recall on it than \ucftwofour,
  as the detector works better thanks to high resolution and quality of 
  the \multisports images. 
  }
  \label{tab:trackers_recall}
 \renewcommand{\arraystretch}{\myarraystretch} 
  \begin{tabular}{lccccc}
      \toprule
       &  
       & \multicolumn{4}{c}{Recall} \\
      &  Fine-tune 
      & \multicolumn{2}{c}{\multisportsshort} & \multicolumn{2}{c}{\ucftwofour} \\
     Tracker & detector 
     & train & val & train & val \\
      \midrule
      \tracker & \xmark 
      & 84.0 & 84.5 & 25.0 & 26.8 \\
      \tracker & \cmark 
      & 93.1 & 91.0 & 94.9 & 84.0 \\
       \bottomrule
  \end{tabular}
  \vskip -3mm
\end{table}

\qheading{Tracker as filtering module:} Using trackers as a post
processing step for action detection has many advantages,
which we demonstrate in all the above tables, including \cref{tab:ucf24},
where we get substantial improvement in \fmap{}, labeled as ``Baseline + tracks''.
Firstly, the tracker helps filter out
false positive person detections with high scores that spuriously appear for a few frames. This
reduces the load on person bounding box thresholding.
Most of the current \sota methods use a relatively high threshold to filter out
unwanted false positive person detections,
e.g. \pyslowfast~\cite{fan2020pyslowfast}
uses $0.8$ and mmaction2~\cite{mmaction2} uses $0.9$.
Yet, such strict thresholds can eliminate some crucial true positives. 
In contrast to standard methods,
we use a relatively liberal ($0.05$) threshold value for our track-based method.
Secondly, using a good tracker greatly simplifies tube construction.
Trackers are specifically designed to solve the linking problem,
removing the need for greedy linking algorithms
used in prior work \cite{singh2017online, kalogeiton2017action,li2020actionsas}. 
The performance gains, both in \vmap{} and \fmap{}, obtained
by ``Baseline + tracks'' rows of \cref{tab:multisports} and \cref{tab:ucf24}
over the ``Baseline'' row, clearly demonstrate this.

\subsection{Discussion}
\label{subsec:discussion}

In this work, our main objective is
to study action detection under large motion.
The experiments on \multisports and \ucftwofour,
see \cref{tab:splits,tab:motion_ap},
demonstrate that \taad, \ie utilizing track information for 
feature aggregation, improves performance across the board.
This does not mean that there is no room for further improvement.
Our method is
sensitive to the performance of the tracker, since this is the first step of our pipeline. 
Using 
a better state-of-the-art tracker and person detector,
such as the ones employed by other contemporary methods \cite{pan2021actor,tang2020asynchronous,ning2021person,kopuklu2019you}),
should boost performance further, especially on \ucftwofour,
where \yolovfive struggles. 
Moreover, we can improve action detection performance by incorporating spatial/actor context modelling~\cite{tuberZhao2022,pan2021actor}, long-term temporal context~\cite{tang2020asynchronous}, or a transformer head~\cite{tuberZhao2022} or backbone~\cite{liu2022video,li2022mvitv2} into \taad.

\begin{figure}[t] 
  \centering
  \captionsetup{font=footnotesize}
  \includegraphics[width=.47\textwidth]{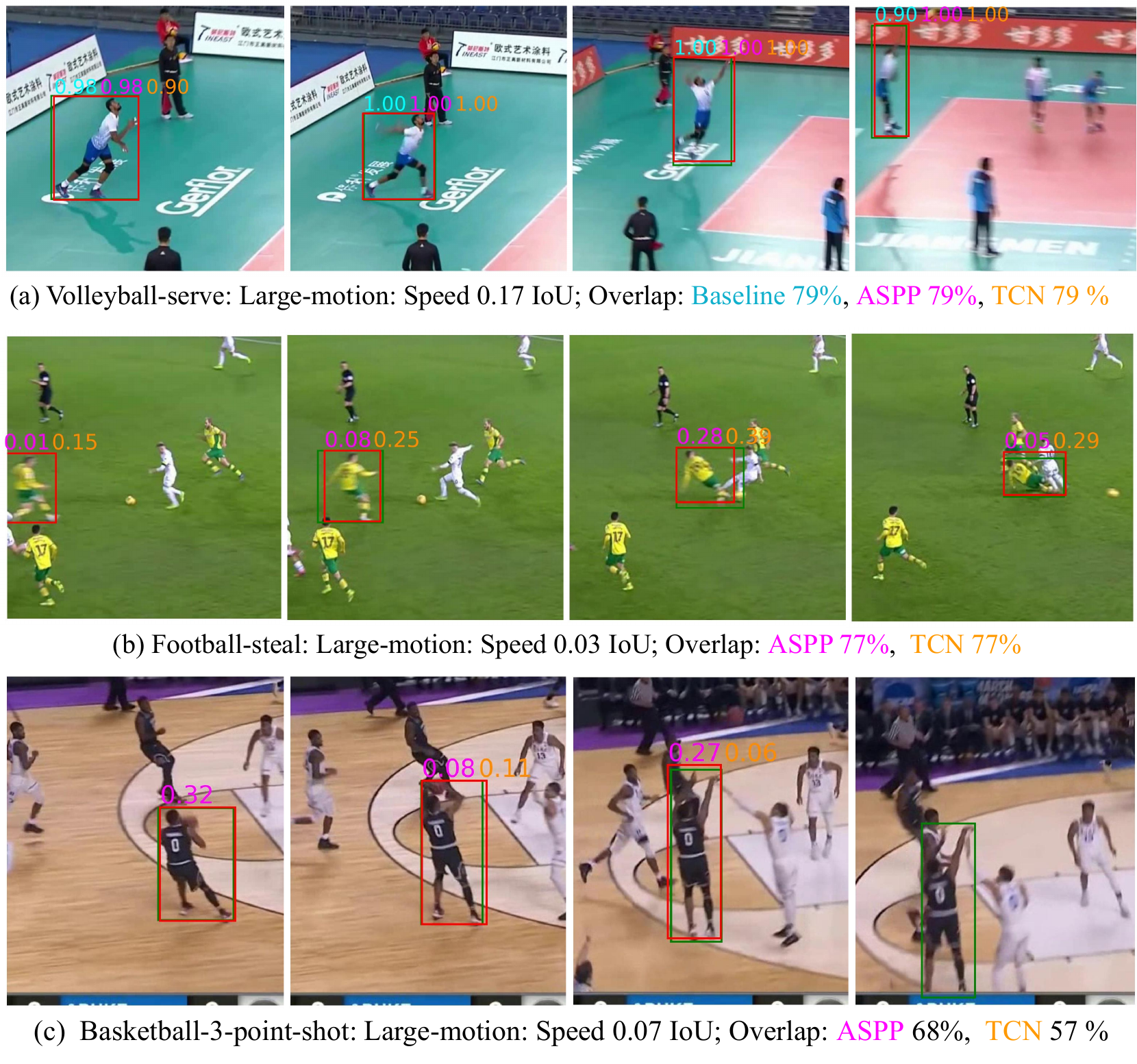}
  \vskip -2mm
  \caption{
  \lmotion due to fast action and camera movement in Volley-spike instance (a)  detected by all the methods including baseline,
  but in (b), Football-steal instance is only detected by ASPP and TCN.
  (c) \lmotion (0.07) due to camera, baseline fails to detect and ASPP module shows better overlap than TCN.} 
  \label{fig:large_motion_visual} 
\end{figure}


One could argue that our definition of motion categories is not precise.
Unlike the object size categories
in \mscoco~\cite{lin2014microsoft},
motion categories are not easy to define.
Apart from the complex camera motion (incl. zoom, translation and rotation), which
is pretty common, and quick actor motion, both of which we
show in \cref{fig:reasons},
special care has to be taken to avoid mislabelling cyclic motions.
\multisports for example contains multiple actions,
\eg in aerobics,
where the actor
starts and ends at the same position. This would result in a high \iou
between the initial and last boxes and thus an erroneous small motion label.
To solve this problem, we use an average of \iou{s} computed
at different frame offsets.
While our motion labelling scheme is not perfect, visual
examples show that it correlates well with motion speed. 
Lastly, Fig. ~\ref{fig:large_motion_visual} shows examples where the baseline fails to detect action tubes but \taad is detects them.
We will provide more qualitative examples
in the \supmat to illustrate this point.
\section{Conclusion}
\label{sec:conclusion}

In this work, we analyse and identify three coarse motion categories
in action detection datasets. We observe
that existing action detection methods struggle in the presence of
large motions, \eg
motion due to fast actor movement or large camera motion,
To remedy this, 
We introduce \taadlong (\taad), a method that utilizes
actor tracks to solve this problem. \taad aggregates information
across actor tracks, rather than using a tube made from proposal boxes.
We evaluate the proposed method on two datasets,
\multisports and \ucftwofour. \multisports is the ideal benchmark
for this task, thanks to its large number of instances
with fast-paced actions. \taad not only bridges the performance
gap between motion categories, but also sets
a new state-of-the-art for \multisports by
beating last year's challenge winner by a large margin. 



\par\vfill\par
\clearpage

{\small
\bibliographystyle{ieee_fullname}
\bibliography{ref}

\begin{thebibliography}{10}\itemsep=-1pt

\bibitem{behl2017incremental}
Harkirat~S. Behl, Michael Sapienza, Gurkirt Singh, Suman Saha, Fabio Cuzzolin,
  and Philip H.~S. Torr.
\newblock Incremental tube construction for human action detection.
\newblock In {\em {British Machine Vision Conference (BMVC)}}, 2018.

\bibitem{yolov5deepsort2020}
Mikel Broström.
\newblock Real-time multi-object tracker using yolov5 and deep sort.
\newblock \url{https://github.com/mikel-brostrom/Yolov5_DeepSort_Pytorch},
  2020.

\bibitem{carion2020end}
Nicolas Carion, Francisco Massa, Gabriel Synnaeve, Nicolas Usunier, Alexander
  Kirillov, and Sergey Zagoruyko.
\newblock End-to-end object detection with transformers.
\newblock In {\em European conference on computer vision}, pages 213--229.
  Springer, 2020.

\bibitem{carreira2017quo}
Joao Carreira and Andrew Zisserman.
\newblock {Quo Vadis, Action Recognition? A New Model and the Kinetics
  Dataset}.
\newblock In {\em {Computer Vision and Pattern Recognition (CVPR)}}, pages
  4724--4733, 2017.

\bibitem{deeplabv3plus2018}
Liang-Chieh Chen, Yukun Zhu, George Papandreou, Florian Schroff, and Hartwig
  Adam.
\newblock Encoder-decoder with atrous separable convolution for semantic image
  segmentation.
\newblock In {\em {European Conference on Computer Vision (ECCV)}}, 2018.

\bibitem{chen2021watch}
Shoufa Chen, Peize Sun, Enze Xie, Chongjian Ge, Jiannan Wu, Lan Ma, Jiajun
  Shen, and Ping Luo.
\newblock {Watch Only Once: An End-to-End Video Action Detection Framework}.
\newblock In {\em {International Conference on Computer Vision ({ICCV})}},
  pages 8178--8187, 2021.

\bibitem{mmaction2}
MMAction2 Contributors.
\newblock Openmmlab's next generation video understanding toolbox and
  benchmark.
\newblock \url{https://github.com/open-mmlab/mmaction2}, 2020.

\bibitem{dave2022gabriellav2}
Ishan Dave, Zacchaeus Scheffer, Akash Kumar, Sarah Shiraz, Yogesh~Singh Rawat,
  and Mubarak Shah.
\newblock Gabriellav2: Towards better generalization in surveillance videos for
  action detection.
\newblock In {\em Proceedings of the IEEE/CVF Winter Conference on Applications
  of Computer Vision}, pages 122--132, 2022.

\bibitem{fan2020pyslowfast}
Haoqi Fan, Yanghao Li, Bo Xiong, Wan-Yen Lo, and Christoph Feichtenhofer.
\newblock Pyslowfast.
\newblock \url{https://github.com/facebookresearch/slowfast}, 2020.

\bibitem{fan2021multiscale}
Haoqi Fan, Bo Xiong, Karttikeya Mangalam, Yanghao Li, Zhicheng Yan, Jitendra
  Malik, and Christoph Feichtenhofer.
\newblock Multiscale vision transformers.
\newblock In {\em {International Conference on Computer Vision ({ICCV})}},
  pages 6824--6835, 2021.

\bibitem{feichtenhofer2020x3d}
Christoph Feichtenhofer.
\newblock {X3D: Expanding Architectures for Efficient Video Recognition}.
\newblock In {\em {Computer Vision and Pattern Recognition (CVPR)}}, pages
  203--213, 2020.

\bibitem{feichtenhofer2019slowfast}
Christoph Feichtenhofer, Haoqi Fan, Jitendra Malik, and Kaiming He.
\newblock {SlowFast networks for video recognition}.
\newblock In {\em {Computer Vision and Pattern Recognition (CVPR)}}, pages
  6202--6211, 2019.

\bibitem{girdhar2018better}
Rohit Girdhar, Jo{\~a}o Carreira, Carl Doersch, and Andrew Zisserman.
\newblock {A Better Baseline for AVA}.
\newblock {\em arXiv preprint arXiv:1807.10066}, 2018.

\bibitem{Georgia-2015a}
Georgia Gkioxari and Jitendra Malik.
\newblock {Finding action tubes}.
\newblock In {\em {Computer Vision and Pattern Recognition (CVPR)}}, pages
  759--768, 2015.

\bibitem{gu2018ava}
Chunhui Gu, Chen Sun, David~A Ross, Carl Vondrick, Caroline Pantofaru, Yeqing
  Li, Sudheendra Vijayanarasimhan, George Toderici, Susanna Ricco, Rahul
  Sukthankar, et~al.
\newblock {AVA: A Video Dataset of Spatio-temporally Localized Atomic Visual
  Actions }.
\newblock In {\em {Computer Vision and Pattern Recognition (CVPR)}}, pages
  6047--6056, 2018.

\bibitem{he2017mask}
Kaiming He, Georgia Gkioxari, Piotr Doll{\'a}r, and Ross Girshick.
\newblock {Mask R-CNN}.
\newblock In {\em {International Conference on Computer Vision ({ICCV})}},
  pages 2961--2969, 2017.

\bibitem{he2016deep}
Kaiming He, Xiangyu Zhang, Shaoqing Ren, and Jian Sun.
\newblock Deep residual learning for image recognition.
\newblock In {\em {Computer Vision and Pattern Recognition (CVPR)}}, pages
  770--778, 2016.

\bibitem{kalogeiton2017action}
Vicky Kalogeiton, Philippe Weinzaepfel, Vittorio Ferrari, and Cordelia Schmid.
\newblock {Action Tubelet Detector for Spatio-Temporal Action Localization}.
\newblock In {\em {International Conference on Computer Vision ({ICCV})}},
  2017.

\bibitem{kopuklu2019you}
Okan K{\"o}p{\"u}kl{\"u}, Xiangyu Wei, and Gerhard Rigoll.
\newblock {You Only Watch Once: A Unified CNN Architecture for Real-Time
  Spatiotemporal Action Localization}.
\newblock {\em arXiv preprint arXiv:1911.06644}, 2019.

\bibitem{li2021multisports}
Yixuan Li, Lei Chen, Runyu He, Zhenzhi Wang, Gangshan Wu, and Limin Wang.
\newblock {MultiSports: A Multi-Person Video Dataset of Spatio-Temporally
  Localized Sports Actions}.
\newblock In {\em {International Conference on Computer Vision ({ICCV})}},
  pages 13536--13545, 2021.

\bibitem{li2020actionsas}
Yixuan Li, Zixu Wang, Limin Wang, and Gangshan Wu.
\newblock Actions as moving points.
\newblock In {\em {European Conference on Computer Vision (ECCV)}}, 2020.

\bibitem{li2022mvitv2}
Yanghao Li, Chao-Yuan Wu, Haoqi Fan, Karttikeya Mangalam, Bo Xiong, Jitendra
  Malik, and Christoph Feichtenhofer.
\newblock Mvitv2: Improved multiscale vision transformers for classification
  and detection.
\newblock In {\em Proceedings of the IEEE/CVF Conference on Computer Vision and
  Pattern Recognition (CVPR)}, pages 4804--4814, June 2022.

\bibitem{lin2014microsoft}
Tsung-Yi Lin, Michael Maire, Serge Belongie, James Hays, Pietro Perona, Deva
  Ramanan, Piotr Doll{\'a}r, and C~Lawrence Zitnick.
\newblock {Microsoft COCO: Common Objects in Context}.
\newblock In {\em {European Conference on Computer Vision (ECCV)}}, pages
  740--755. Springer, 2014.

\bibitem{liu2022convnet}
Zhuang Liu, Hanzi Mao, Chao-Yuan Wu, Christoph Feichtenhofer, Trevor Darrell,
  and Saining Xie.
\newblock {A ConvNet for the 2020s}.
\newblock {\em {Computer Vision and Pattern Recognition (CVPR)}}, 2022.

\bibitem{liu2022video}
Ze Liu, Jia Ning, Yue Cao, Yixuan Wei, Zheng Zhang, Stephen Lin, and Han Hu.
\newblock Video swin transformer.
\newblock In {\em Proceedings of the IEEE/CVF Conference on Computer Vision and
  Pattern Recognition}, pages 3202--3211, 2022.

\bibitem{cosine_lr}
Ilya Loshchilov and Frank Hutter.
\newblock {SGDR:} stochastic gradient descent with warm restarts.
\newblock In {\em 5th International Conference on Learning Representations,
  {ICLR} 2017, Toulon, France, April 24-26, 2017, Conference Track
  Proceedings}. OpenReview.net, 2017.

\bibitem{ning2021person}
Zhiqing Ning, Qiaokang Xie, Wengang Zhou, Liangwei Wang, and Houqiang Li.
\newblock {Person-Context Cross Attention for Spatio-Temporal Action
  Detection}.
\newblock Technical report, Huawei Noah's Ark Lab, and University of Science
  and Technology of China, 2021.

\bibitem{pan2021actor}
Junting Pan, Siyu Chen, Mike~Zheng Shou, Yu Liu, Jing Shao, and Hongsheng Li.
\newblock {Actor-Context-Actor Relation Network for Spatio-Temporal Action
  Localization}.
\newblock In {\em {Computer Vision and Pattern Recognition (CVPR)}}, pages
  464--474, 2021.

\bibitem{peng2016eccv}
Xiaojiang Peng and Cordelia Schmid.
\newblock {Multi-region two-stream R-CNN for action detection}.
\newblock In {\em {European Conference on Computer Vision (ECCV)}}, pages
  744--759, 2016.

\bibitem{redmon2016you}
Joseph Redmon, Santosh Divvala, Ross Girshick, and Ali Farhadi.
\newblock {You Only Look Once: Unified, Real-Time Object Detection }.
\newblock In {\em {Computer Vision and Pattern Recognition (CVPR)}}, pages
  779--788, 2016.

\bibitem{saha2017amtnet}
Suman Saha, Gurkirt Singh, and Fabio Cuzzolin.
\newblock {AMTnet: Action-Micro-Tube regression by end-to-end trainable deep
  architecture}.
\newblock In {\em {International Conference on Computer Vision ({ICCV})}},
  2017.

\bibitem{saha2016deep}
Suman Saha, Gurkirt Singh, Michael Sapienza, Philip~HS Torr, and Fabio
  Cuzzolin.
\newblock Deep learning for detecting multiple space-time action tubes in
  videos.
\newblock In {\em {British Machine Vision Conference (BMVC)}}, 2016.

\bibitem{singh2022road}
Gurkirt Singh, Stephen Akrigg, Manuele Di~Maio, Valentina Fontana,
  Reza~Javanmard Alitappeh, Suman Saha, Kossar Jeddisaravi, Farzad Yousefi,
  Jacob Culley, Tom Nicholson, et~al.
\newblock Road: The road event awareness dataset for autonomous driving.
\newblock {\em {Transactions on Pattern Analysis and Machine Intelligence
  (TPAMI)}}, 1(01):1--1, feb 5555.

\bibitem{singh2019recurrent}
Gurkirt SingH and Fabio Cuzzolin.
\newblock Recurrent convolutions for causal 3d cnns.
\newblock In {\em Proceedings of the IEEE/CVF International Conference on
  Computer Vision Workshops}, pages 0--0, 2019.

\bibitem{singh2018predicting}
Gurkirt Singh, Suman Saha, and Fabio Cuzzolin.
\newblock Predicting action tubes.
\newblock In {\em Proceedings of the European Conference on Computer Vision
  (ECCV) Workshops}, pages 0--0, 2018.

\bibitem{singh2018tramnet}
Gurkirt Singh, Suman Saha, and Fabio Cuzzolin.
\newblock {TraMNet-Transition Matrix Network for Efficient Action Tube
  Proposals}.
\newblock In {\em {Asian Conference on Computer Vision (ACCV)}}, pages
  420--437. Springer, 2018.

\bibitem{singh2017online}
Gurkirt Singh, Suman Saha, Michael Sapienza, Philip~HS Torr, and Fabio
  Cuzzolin.
\newblock {Online Real-time Multiple Spatiotemporal Action Localisation and
  Prediction}.
\newblock In {\em {International Conference on Computer Vision ({ICCV})}},
  pages 3637--3646, 2017.

\bibitem{song2019tacnet}
Lin Song, Shiwei Zhang, Gang Yu, and Hongbin Sun.
\newblock {TACNet: Transition-aware context network for spatio-temporal action
  detection}.
\newblock In {\em {Computer Vision and Pattern Recognition (CVPR)}}, pages
  11987--11995, 2019.

\bibitem{soomro2016predicting}
Khurram Soomro, Haroon Idrees, and Mubarak Shah.
\newblock Predicting the where and what of actors and actions through online
  action localization.
\newblock In {\em {Computer Vision and Pattern Recognition (CVPR)}}, pages
  2648--2657, 2016.

\bibitem{soomro2012ucf101}
Khurram Soomro, Amir~Roshan Zamir, and Mubarak Shah.
\newblock {UCF101: A Dataset of 101 Human Actions Classes From Videos in The
  Wild}, 2012.

\bibitem{tang2020asynchronous}
Jiajun Tang, Jin Xia, Xinzhi Mu, Bo Pang, and Cewu Lu.
\newblock Asynchronous interaction aggregation for action detection.
\newblock In {\em {European Conference on Computer Vision (ECCV)}}, pages
  71--87. Springer, 2020.

\bibitem{yolov5}
ultralytics.
\newblock Yolov5: Real-time object detector.
\newblock \url{https://ultralytics.com/yolov5}, 2020.

\bibitem{van2015apt}
Jan~C Van~Gemert, Mihir Jain, Ella Gati, Cees~GM Snoek, et~al.
\newblock {APT: Action localization proposals from dense trajectories}.
\newblock In {\em {British Machine Vision Conference (BMVC)}}, volume~2,
  page~4, 2015.

\bibitem{wang2018nonlocal}
Xiaolong Wang, Ross Girshick, Abhinav Gupta, and Kaiming He.
\newblock {Non-local Neural Networks}.
\newblock In {\em {Computer Vision and Pattern Recognition (CVPR)}}, pages
  7794--7803, 2018.

\bibitem{weinzaepfel2015learning}
Philippe Weinzaepfel, Zaid Harchaoui, and Cordelia Schmid.
\newblock Learning to track for spatio-temporal action localization.
\newblock In {\em {International Conference on Computer Vision ({ICCV})}},
  pages 3164--3172, 2015.

\bibitem{weinzaepfel2021mimetics}
Philippe Weinzaepfel and Gr{\'e}gory Rogez.
\newblock Mimetics: Towards understanding human actions out of context.
\newblock {\em International Journal of Computer Vision}, 129(5):1675--1690,
  2021.

\bibitem{wu2019long}
Chao-Yuan Wu, Christoph Feichtenhofer, Haoqi Fan, Kaiming He, Philipp
  Krahenbuhl, and Ross Girshick.
\newblock {Long-Term Feature Banks for Detailed Video Understanding }.
\newblock In {\em {Computer Vision and Pattern Recognition (CVPR)}}, pages
  284--293, 2019.

\bibitem{wu2019detectron2}
Yuxin Wu, Alexander Kirillov, Francisco Massa, Wan-Yen Lo, and Ross Girshick.
\newblock Detectron2.
\newblock \url{https://github.com/facebookresearch/detectron2}, 2019.

\bibitem{yang2019step}
Xitong Yang, Xiaodong Yang, Ming-Yu Liu, Fanyi Xiao, Larry~S Davis, and Jan
  Kautz.
\newblock {STEP: Spatio-Temporal Progressive Learning for Video Action
  Detection}.
\newblock In {\em {Computer Vision and Pattern Recognition (CVPR)}}, pages
  264--272, 2019.

\bibitem{yu2018deep}
Fisher Yu, Dequan Wang, Evan Shelhamer, and Trevor Darrell.
\newblock Deep layer aggregation.
\newblock In {\em Proceedings of the IEEE conference on computer vision and
  pattern recognition}, pages 2403--2412, 2018.

\bibitem{zhang2019structured}
Yubo Zhang, Pavel Tokmakov, Martial Hebert, and Cordelia Schmid.
\newblock A structured model for action detection.
\newblock In {\em {Computer Vision and Pattern Recognition (CVPR)}}, pages
  9975--9984, 2019.

\bibitem{zhao2019dance}
Jiaojiao Zhao and Cees~GM Snoek.
\newblock {Dance with Flow: Two-in-One Stream Action Detection }.
\newblock In {\em {Computer Vision and Pattern Recognition (CVPR)}}, pages
  9935--9944, 2019.

\bibitem{tuberZhao2022}
Jiaojiao Zhao, Yanyi Zhang, Xinyu Li, Hao Chen, Bing Shuai, Mingze Xu, Chunhui
  Liu, Kaustav Kundu, Yuanjun Xiong, Davide Modolo, Ivan Marsic, Cees G.~M.
  Snoek, and Joseph Tighe.
\newblock Tuber: Tubelet transformer for video action detection.
\newblock In {\em Proceedings of the IEEE/CVF Conference on Computer Vision and
  Pattern Recognition (CVPR)}, pages 13598--13607, June 2022.

\bibitem{torchreid}
Kaiyang Zhou and Tao Xiang.
\newblock Torchreid: A library for deep learning person re-identification in
  pytorch.
\newblock {\em arXiv preprint arXiv:1910.10093}, 2019.

\bibitem{zhou2021osnet}
Kaiyang Zhou, Yongxin Yang, Andrea Cavallaro, and Tao Xiang.
\newblock {Learning Generalisable Omni-Scale Representations for Person
  Re-Identification}.
\newblock {\em {Transactions on Pattern Analysis and Machine Intelligence
  (TPAMI)}}, 2021.

\end{thebibliography}
}

\end{document}


\pagestyle{headings}
\mainmatter
\def\ECCVSubNumber{2028}  

\title{\ourTitleSupMat} 

\titlerunning{ECCV-22 submission ID \ECCVSubNumber} 
\authorrunning{ECCV-22 submission ID \ECCVSubNumber} 
\author{Anonymous ECCV submission}
\institute{Paper ID \ECCVSubNumber}

\maketitle
\section{Overview}
\label{experiments}

\begin{figure*}[t] 
  \centering
  \includegraphics[scale=0.5]{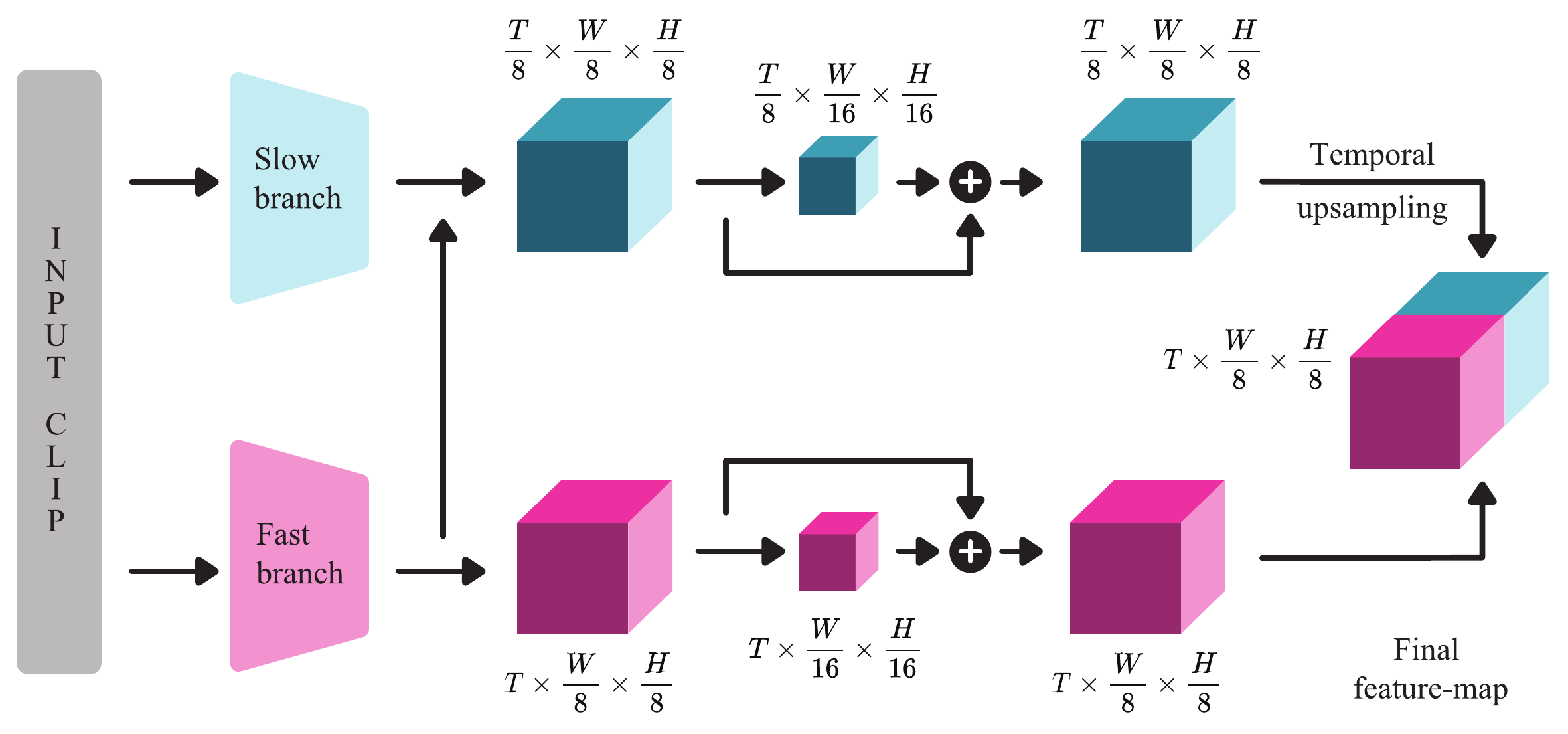}
  \caption{
  Single backbone with a single spatial upsample/downward step from ${res}_{5}$ to ${res}_{4}$. We add a single feature pyramid network (FPN) block to increase the spatial resolution,
  because the average size ($26\times54$) of bounding boxes is very small,
  compared to the size ($256\times256$) of the input image fed to network,
  \eg when using \multisports data.
  }
  \label{fig:backbone} 
\end{figure*}

Here, we aim to provide additional details about the certain parts of main paper.
First, we show architecture of backbone network, where feature pyramid network structure (FPN) incorporated into  \slowfast~\cite{feichtenhofer2019slowfast} network in figure~\reffig{fig:backbone}. 
Second, we show the structure of \tfa modules used in our \taad model in \refsec{sec:tfa_structure}. 
Then, we present frame-level \motionmap and \motionap on individual time scales in \refsec{sec:indiv_scale} 
Finally, we show visual results of detected action-tube instances under different motion types in \refsec{subsec:visuals}.

\section{Structure of \tfalong Modules}
\label{sec:tfa_structure}

\refListing{lst:maxpool}, ~\ref{lst:tcn},
and ~\ref{lst:aspp} contain the \pytorch implementation
of our MaxPool, \tcn and \aspp \tfa modules,
used with our \taadlong (\taad)
method.
Similar to 1D-\aspp module (see Listing-Fig.~\ref{lst:aspp}),
In addition to these blocks, 
we tried
1D-ConvNeXt~\cite{liu2022convnet} and 1D-\swin~\cite{liu2022video} blocks as well,
but observed that training was unstable and that the final performance was worse.
For example,
1D-ConvNext could only reach up to 44\% \fmap{} compared to 53.3\% using MaxPool module (Listing-Fig.~\ref{lst:maxpool}).
In all the listings that follow, $C$ is the number of channels and $T$ is the number 
of input frames.

\begin{figure*}
\begin{lstlisting}[caption={MaxPool module with input feature of size $T \times C$ and output of $1 \times C$.},captionpos=b,label={lst:maxpool}]
    
    {
    (tube_temporal_pool): AdaptiveMaxPool1d(output_size=1)
    }
\end{lstlisting}
\end{figure*}

\begin{figure*}
\begin{lstlisting}[caption={TCN module with input feature of size $T \times C$ with output of $1 \times C$.},captionpos=b,label={lst:tcn}]
    
    {
    (TCN): Conv1d(576, 576, kernel_size=3, 
            stride=1, padding=2, dilation=2)
    (tube_temporal_pool): AdaptiveMaxPool1d(output_size=1)
    }
    
\end{lstlisting}
\end{figure*}

\begin{figure*}
\begin{lstlisting}[caption={1D-ASPP~\cite{deeplabv3plus2018} module with input feature of size $T \times C$ with output of $1 \times C$.}, captionpos=b, label={lst:aspp}]
    
    {
    (ASPP): ASPP1D(
        (convs): ModuleList(
          (0): Conv1d(576, 256, kernel_size=1, 
                      stride=1)
          (1): Sequential(
            (0): Conv1d(256, 576, kernel_size=1, 
                        stride=1)
            (1): ReLU()
          )
          2: ASPPConv1D(
            (0): Conv1d(256, 576, kernel_size=3, 
                stride=1, padding=1)
            (1): ReLU()
          )
          (3): ASPPConv1D(
            (0): Conv1d(256, 576, kernel_size=3, 
                stride=1, padding=3, dilation=3)
            (1): ReLU()
          )
          (4): ASPPConv1D(
            (0): Conv1d(256, 576, kernel_size=3, 
                stride=1, padding=(5), dilation=5)
            (1): ReLU()
          )
          (5): ASPPPooling1D(
            (0): AdaptiveAvgPool1d(output_size=1)
            (1): Conv1d(256, 576, kernel_size=1, 
                stride=1)
            2: ReLU()
          )
        )
        (project): Sequential(
          (0): Conv1d(2880, 576, kernel_size=1, 
                        stride=1, bias=False)
          (1): ReLU()
          )
    )
    
    (tube_temporal_pool): AdaptiveMaxPool1d(output_size=1)
    
    }
    
\end{lstlisting}
\end{figure*}

\section{Individual time scales results}
\label{sec:indiv_scale}
Frame-level \motionmap and \motionap{} on individual time scales are shown in 
following tables:
\begin{enumerate}[label=(\arabic*)]
    \item \motionap{} on \multisports in \reftab{tab:motion_ap_ms}
    \item \motionap{} on \ucftwofour in \reftab{tab:motion_ap_ucf24}
    \item \motionmap{} on \multisports in \reftab{tab:motion_map_ms}
    \item \motionmap{} on \ucftwofour in \reftab{tab:motion_map_ucf24}
\end{enumerate}

\begin{table}[t!]
    \centering
    \footnotesize
    \renewcommand{\arraystretch}{\myarraystretch}
    \setlength{\tabcolsep}{8pt}
    
    \caption{
    \motionap{} ablation on \multisports~\cite{li2021multisports}. We investigate the effect of different feature aggregation modules
    using frame \motionap{} to asses the quality of motion-wise action detection.
    Aggregating features across tracks, instead of cuboids,
    improves action detection performance across all categories, with a particularly noticeable
    improvement for large motions.
    }
    \begin{tabular}{lllll}
    \toprule
     & \fmap{} & \multicolumn{3}{c}{\motionap{}} \\
    Method & @0.5   &  Large & Medium & Small \\
    \midrule
     \multicolumn{5}{c}{Speed-\iou measured as mean over time scales [4,8,16,24,26]} \\ 
     \midrule
    Baseline & 49.6 & 63.2 & 77.7 & 82.4   \\
    Baseline + track${}^{*}$ & 50.6 & 64.6 & 78.7 & 84.4  \\ 
    \taad+\maxpool & 53.9  & 70.2 & \textbf{83.4} & 86.1  \\
    \taad+\aspp & 54.4 & \textbf{71.1} & \textbf{83.4} & 86.9  \\
    \taad+\tcn & \textbf{55.3} & 70.4 & 83.3 & \textbf{87.3}\\
       \midrule
     \multicolumn{5}{c}{Speed-\iou measured at time scales of 16 frames} \\ 
     \midrule
    Baseline & 49.6 & 62.7 & 78.8 & 81.8   \\
    Baseline + track${}^{*}$ & 50.6 & 64.2 & 79.8 & 83.7  \\ 
    \taad+\maxpool & 53.9  &  69.9 & 83.9 & 85.9 \\
    \taad+\aspp & 54.4  &  \textbf{70.8} & 84.3 & 86.2  \\
    \taad+\tcn & \textbf{55.3} & 70.0 & \textbf{84.5} & \textbf{86.4} \\
       \midrule
     \multicolumn{5}{c}{Speed-\iou measured at time scales of 24 frame} \\ 
     \midrule
    Baseline & 49.6 &  61.9 & 78.7 & 82.9 \\
    Baseline + track${}^{*}$ & 50.6&  63.3 & 79.7 & 85.0  \\ 
    \taad+\maxpool & 53.9  & 68.6 & \textbf{83.9} & 87.3  \\
    \taad+\aspp & 54.4 &  \textbf{69.6} & 83.5 & \textbf{88.3}   \\
    \taad+\tcn & \textbf{55.3} & 69.0 & 83.7 & \textbf{88.3} \\
    \bottomrule
    \end{tabular}
    \label{tab:motion_ap_ms} 
\end{table}
\begin{table}[t!]
    \centering
    \footnotesize
    \renewcommand{\arraystretch}{\myarraystretch}
    \setlength{\tabcolsep}{8pt}
    
    \caption{
    \motionap{} ablation on \ucftwofour~\cite{soomro2012ucf101}. We investigate the effect of different feature aggregation modules
    using frame \motionap{} to asses the quality of motion-wise action detection.
    Aggregating features across tracks, instead of cuboids,
    improves action detection performance across all categories, with a particularly noticeable
    improvement for large motions.
    }
    \begin{tabular}{lllll}
    \toprule
     & \fmap{} & \multicolumn{3}{c}{\motionap{}} \\
    Method & @0.5   &  Large & Medium & Small \\
    \midrule
     \multicolumn{5}{c}{Speed-\iou measured as mean over time scales [4,8,16,24,26]]} \\ 
     \midrule
     Baseline & 75.9 & 78.8 & 86.5 & 88.5   \\
    Baseline + track${}^{*}$ & 78.3 & 81.4 & 87.8 & 88.4  \\ 
    \taad+\tcn & \textbf{81.5}  & \textbf{82.5} & \textbf{88.7} & \textbf{90.0}\\
       \midrule
     \multicolumn{5}{c}{Speed-\iou measured at time scales of 16 frames} \\ 
     \midrule
     Baseline & 75.9 & 79.0 & 86.7 & 88.2   \\
    Baseline + track${}^{*}$ & 78.3 & 81.4 & 88.3 & 87.8 \\ 
    \taad+\tcn & \textbf{81.5}  & \textbf{82.7} & \textbf{89.0} & \textbf{89.5} \\
       \midrule
     \multicolumn{5}{c}{Speed-\iou measured at time scales of 24 frame} \\ 
     \midrule
     Baseline & 75.9 & 79.0 & 86.3 & 88.7   \\
    Baseline + track${}^{*}$ & 78.3 & 80.9 & 87.9 & 88.6  \\ 
    \taad+\tcn & \textbf{81.5}  & \textbf{82.1} & \textbf{88.8} & \textbf{90.2}   \\
    \bottomrule
    \end{tabular}
    \label{tab:motion_ap_ucf24} 
\end{table}
\begin{table}[t!]
    \centering
    \footnotesize
    \renewcommand{\arraystretch}{\myarraystretch}
    \setlength{\tabcolsep}{8pt}
    
    \caption{
    \motionmap{} ablation on \ucftwofour~\cite{soomro2012ucf101}. We investigate the effect of different feature aggregation modules
    using frame \motionmap{} to asses the quality of motion-wise action detection.
    Aggregating features across tracks, instead of cuboids,
    improves action detection performance across all categories, with a particularly noticeable
    improvement for large motions.
    }
    \begin{tabular}{lllll}
    \toprule
     & \fmap{} & \multicolumn{3}{c}{\motionmap{}} \\
    Method & @0.5   &  Large & Medium & Small \\
    \midrule
     \multicolumn{5}{c}{Speed-\iou measured as mean over time scales [4,8,16,24,26]} \\ 
     \midrule
     Baseline & 75.9 & 67.0 & 77.3 & 70.6  \\
    Baseline + track${}^{*}$ & 78.3 & 68.6 & 79.0 & 72.1  \\ 
    \taad+\tcn & \textbf{81.5}  & \textbf{74.9} & \textbf{83.7} & \textbf{75.1} \\
       \midrule
     \multicolumn{5}{c}{Speed-\iou measured at time scales of 16 frames} \\ 
     \midrule
     Baseline & 75.9 & 68.9 & 78.6 & 72.1   \\
    Baseline + track${}^{*}$ & 78.3 & 70.6 & 80.0 & 73.5 \\ 
    \taad+\tcn & \textbf{81.5}  & \textbf{76.1} & \textbf{84.2} & \textbf{78.1} \\
       \midrule
     \multicolumn{5}{c}{Speed-\iou measured at time scales of 24 frame} \\ 
     \midrule
     Baseline & 75.9 & 67.4 & 78.3 & 72.1  \\
    Baseline + track${}^{*}$ & 78.3 & 68.4 & 79.7 & 74.7  \\ 
    \taad+\tcn & \textbf{81.5}  & \textbf{73.6} & \textbf{83.9} & \textbf{79.1}   \\
    \bottomrule
    \end{tabular}
    \label{tab:motion_map_ucf24} 
\end{table}
\begin{table}[t!]
    \centering
    \footnotesize
    \renewcommand{\arraystretch}{\myarraystretch}
    \setlength{\tabcolsep}{8pt}
    
    \caption{
    \motionmap{} ablation on \multisports~\cite{li2021multisports}. We investigate the effect of different feature aggregation modules
    using frame \motionmap{} to asses the quality of motion-wise action detection.
    Aggregating features across tracks, instead of cuboids,
    improves action detection performance across all categories, with a particularly noticeable
    improvement for large motions.
    In ``Baseline+track\textsuperscript{*}'',
    the track boxes are scored using baseline, with tracks acting as a false-positive filtering mechanism.
    }
    \begin{tabular}{lllll}
    \toprule
     & \fmap{} & \multicolumn{3}{c}{\motionmap{}} \\
    Method & @0.5   &  Large & Medium & Small \\
    \midrule
     \multicolumn{5}{c}{Speed-\iou measured as mean over time scales [4,8,16,24,26]} \\ 
     \midrule
    Baseline & 49.6 & 36.5 & 49.5 & 54.9 \\
    Baseline + track${}^{*}$ & 50.6 & 39.7 & 50.1 & 56.3  \\ 
    \taad + \maxpool & 53.9  & 43.8 & 52.7 & 57.7 \\
    \taad + \aspp & 54.4  & 44.2 & 52.9 & 58.4  \\
    \taad + \tcn & \textbf{55.3} & \textbf{44.9} & \textbf{53.4} & \textbf{60.4}\\
       \midrule
     \multicolumn{5}{c}{Speed-\iou measured at time scales of 16 frames} \\ 
     \midrule
    Baseline & 49.6 & 36.4 & 51.7 & 52.8   \\
    Baseline + track${}^{*}$ & 50.6 & 39.5 & 52.5 & 55.3  \\ 
    \taad+\maxpool & 53.9  & 42.4 & 54.4 & 56.4 \\
    \taad+\aspp & 54.4  &  \textbf{43.7} & 54.2 & 56.3  \\
    \taad+\tcn & \textbf{55.3} & 43.2 & \textbf{55.6} & \textbf{58.9} \\
    \midrule
        \multicolumn{5}{c}{Speed-\iou measured at time scales of 24 frame} \\ 
    \midrule
    Baseline & 49.6 &  32.4 & 51.4 & 55.8 \\
    Baseline + track${}^{*}$ & 50.6 &  35.7 & 51.6 & 57.5  \\ 
    \taad+\maxpool & 53.9  & 38.4 & 54.2 & 58.7  \\
    \taad+\aspp & 54.4 &  39.0 & 53.9 & 59.2   \\
    \taad+\tcn & \textbf{55.3} & \textbf{39.2} & \textbf{54.7} & \textbf{61.0} \\
    \bottomrule
    \end{tabular}
    \label{tab:motion_map_ms} 
\end{table}

The main take-away from these tables is that results are consistent across different time scales
compared to the average over all time scales. 
Through the \motionap{} metric,
we show that
that large-motion action instances are harder to detect compared to medium motions,
which in turn are even harder to detect compared to small-motion action instances, or in other
words performance of large-motion $<$ medium-motion $<$ small-motion.
This result is consistent across both our benchmarks, \ie
\multisports and \ucftwofour.
Such pattern is desirable and intuitive to understand, it is missing in \motionmap{} because some class might not have any (or very few) ground-truth instances in one or two motion type categories resulting very small value of \map or zero \map, since medium is middle motion category it has more classes with some instances with medium motion class, hence fewer classes with zero \motionap{} resulting in higher mean-AP i.e. \motionmap{} for medium motion type.


\section{Motion-wise visual results}
\label{subsec:visuals}
\begin{figure*}[ht] 
  \centering
  \includegraphics[width=1.0\textwidth]{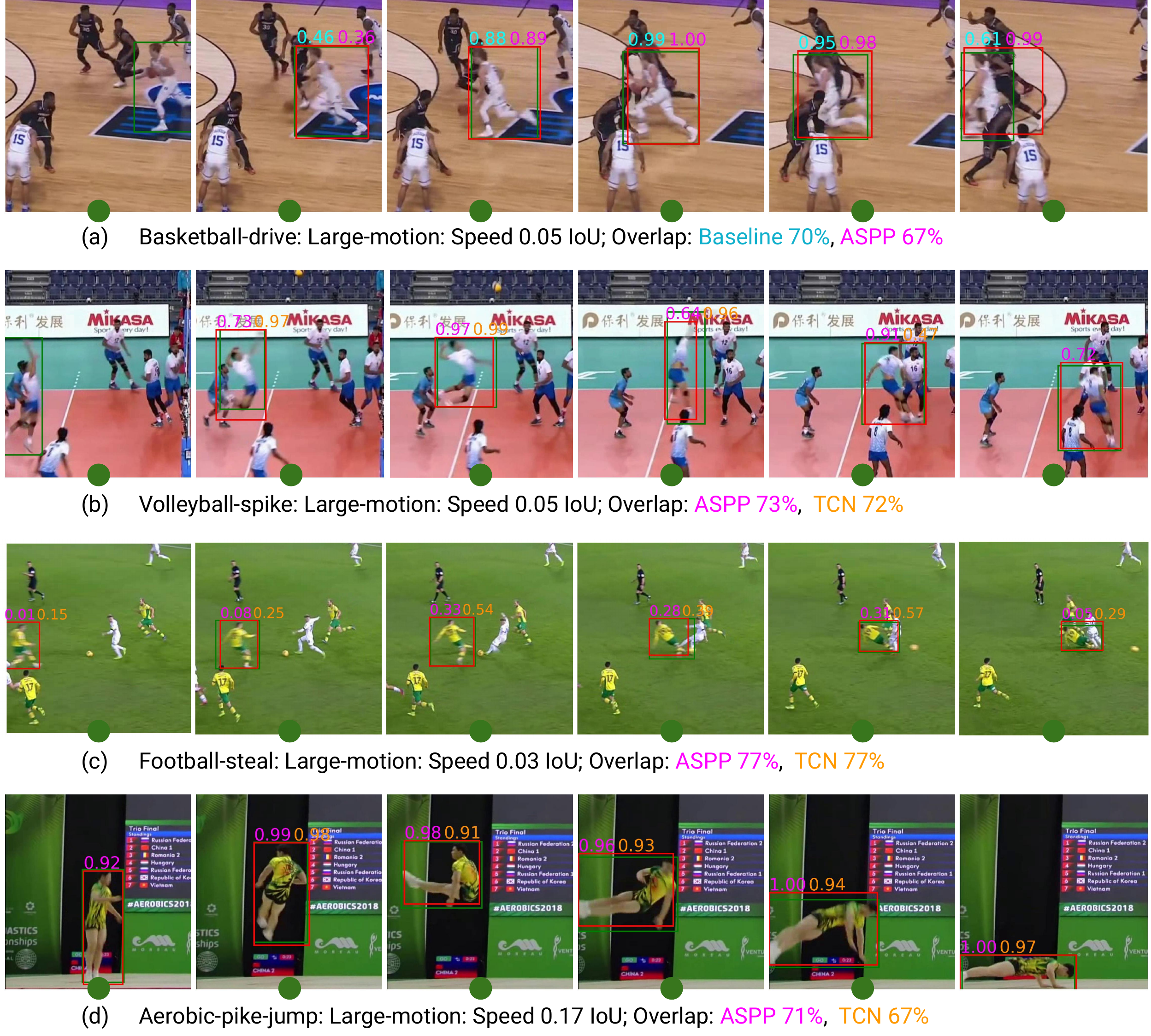}
  \caption{\lmotion due to fast action;
  (a) TCN fail to detect it and others fail to detect initial few frames.
  (b) Volley-spike instance detected with high overlap both by \aspp and \tcn. Similarly, baseline fails to detect fast action instances in (c) and (d). (d) connect back to instances shown in Fig 2 (c) in the introduction Section of main paper.} 
  \label{fig:large_motion_action} 
\end{figure*}

\begin{figure*}[ht] 
  \centering
  \includegraphics[width=1.0\textwidth]{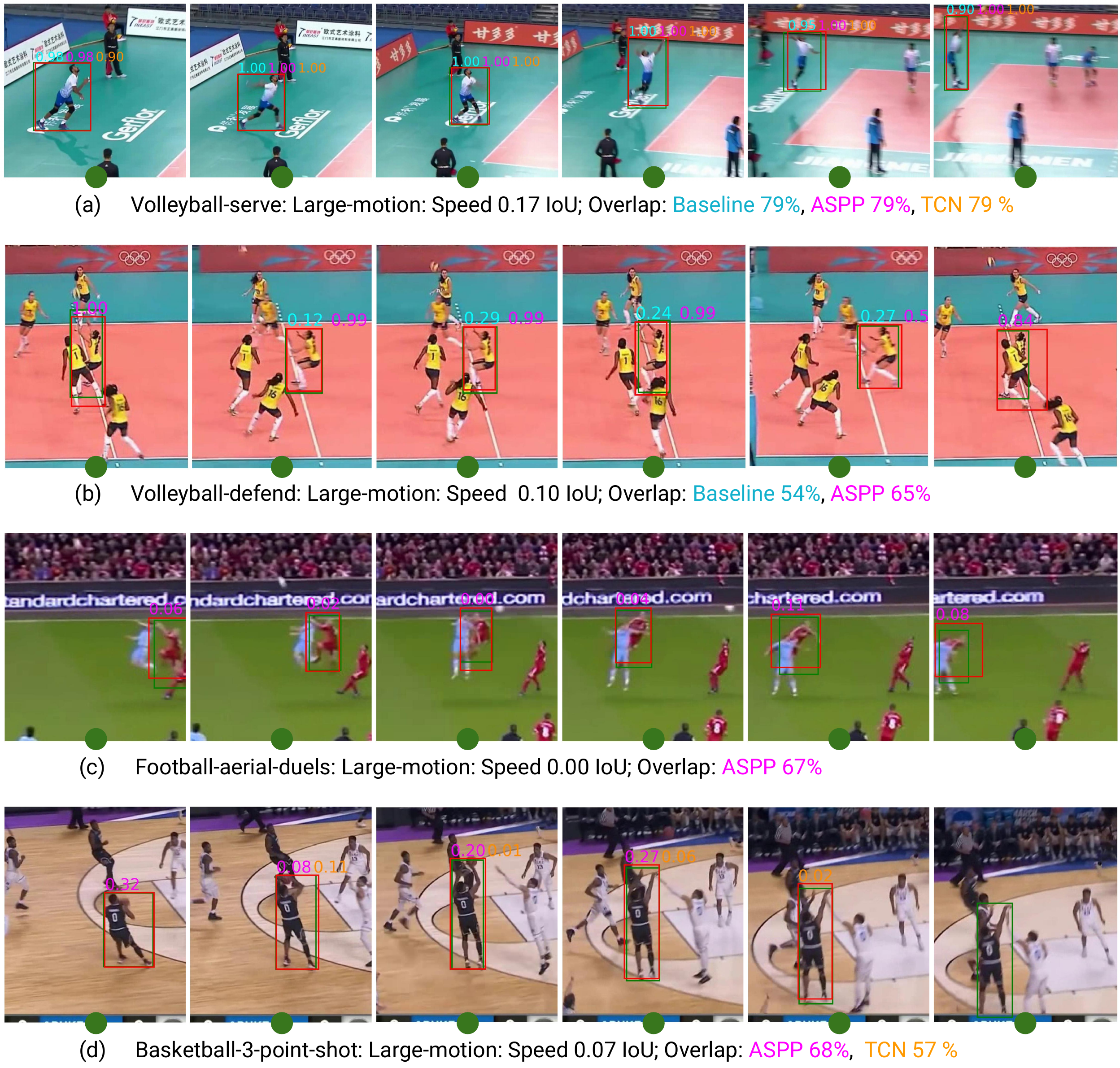}
  \caption{\lmotion due to camera motion;
  (a) shows an instance of ``Volleyball-serve'' which is correctly detected by all three methods, it happens quite often.
  In contrast, \aspp is better at detection large motion instances as shown in (b), (c) and (d). 
  } 
  \label{fig:large_motion_camera_plus} 
\end{figure*}

\begin{figure*}[ht] 
  \centering
  \includegraphics[width=1.0\textwidth]{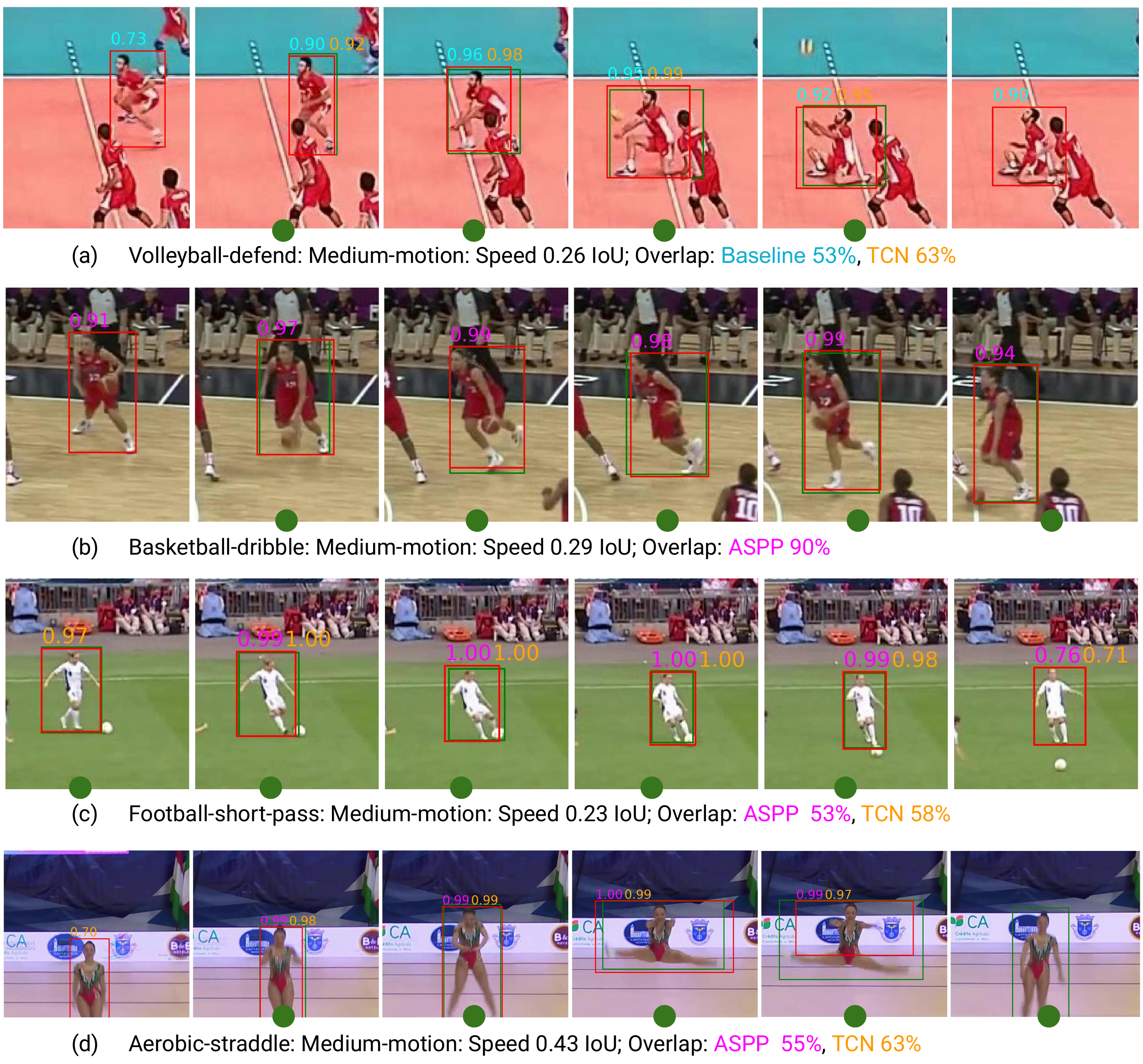}
  \caption{\mmotion;
    all these instances show where temporal detection bounding is longer than ground truth tube, which happen often for medium-motion instances. Accuracy of temporal boundary detection is directly proportional to stability in continuous scores of consecutive boxes in a track, where \aspp seems to be better in these examples as a result having higher overlap in (b) and (c).}
  \label{fig:medium_motion} 
\end{figure*}

\begin{figure*}[ht] 
  \centering
  \includegraphics[width=1.0\textwidth]{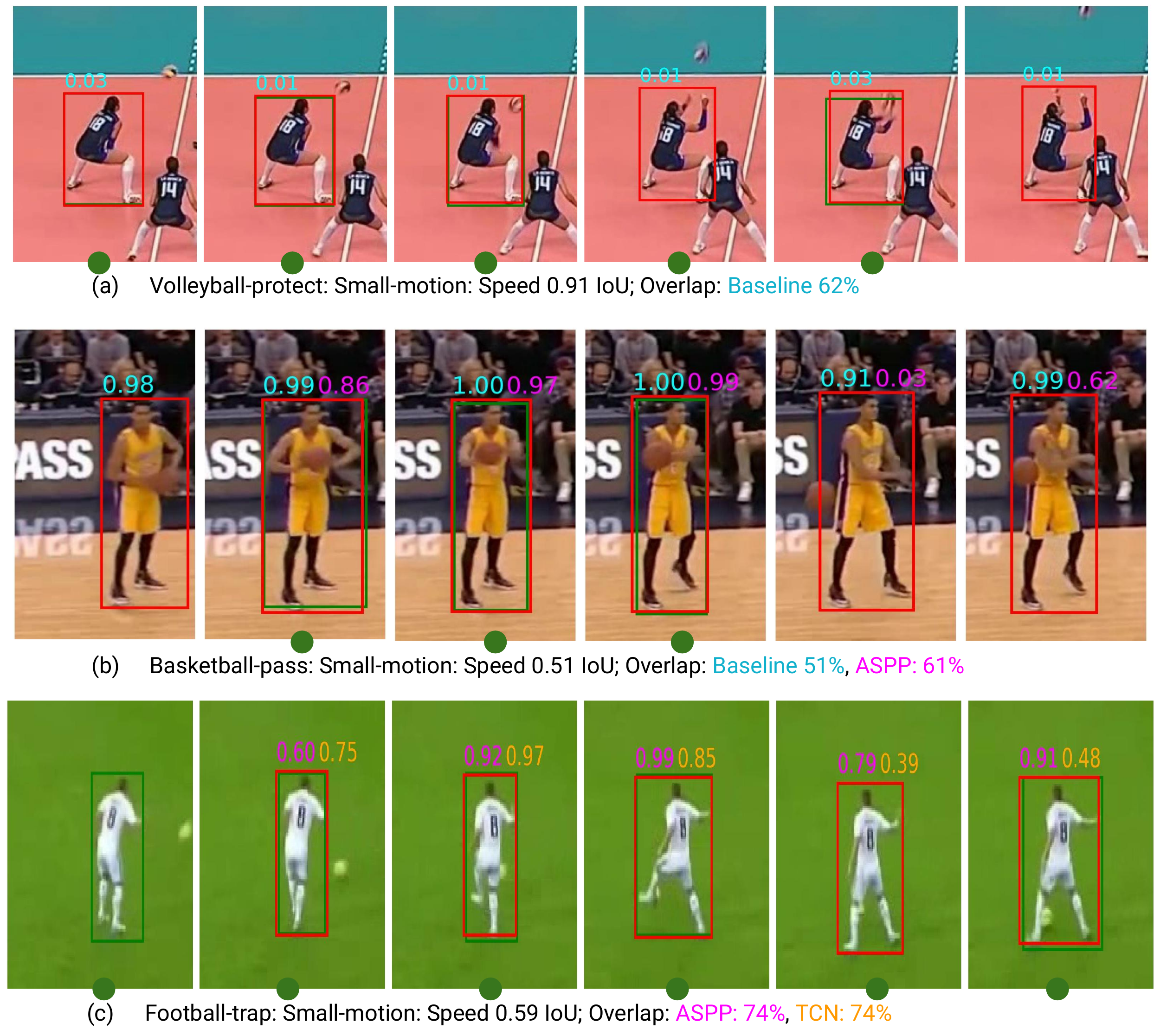}
  \caption{\smotion;
  SubFig. (a) show instances where baseline is able to detect an instance of ``Volleyball-protect'',
  where others not, even though the confidence of the detection is very low for the baseline.
  (b) shows instances where \taad+\tcn fails to detect it.
  (a) shows an action instance where baseline fails, although other methods are able to detect it with high overlap but not in initial few frames.} 
  \label{fig:small_motion} 
\end{figure*}

\begin{figure*}[ht] 
  \centering
  \includegraphics[width=1.0\textwidth]{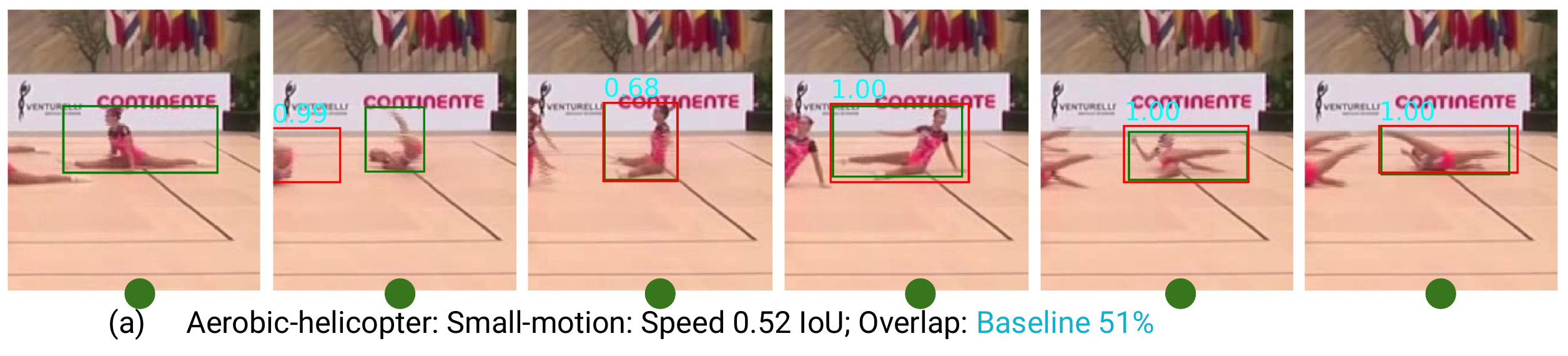}
  \caption{Tracking-error;
  This figures show that if tracking fails, \eg in the first two frames of this action instance, \taad fails to detect such instances.
  } 
  \label{fig:tracking_error} 
\end{figure*}

In this section we show visual results obtained using our baseline and \taad model.
We discuss some interesting observation in the caption of figures. The figures are
best viewed in colour. The qualitative results contain the following scenarios:
\begin{enumerate}[label=(\arabic*)]
    \item \lmotion due to fast execution of actions in \reffig{fig:large_motion_action}.
    \item \lmotion due to fast camera motion in \reffig{fig:large_motion_camera_plus}.
    \item \mmotion action instances in \reffig{fig:medium_motion}.
    \item \smotion action instance in \reffig{fig:small_motion}.
    \item An action instance where \taad fails due to tracking error shown in \reffig{fig:tracking_error}.
\end{enumerate}

In all the captions, 
``Overlap'' denotes the spatiotemporal overlap of the detected tube with the ground-truth tube,
as defined by Weinzaepfel \etal \cite{weinzaepfel2015learning}.
Ground-truth boxes and frames (dot at the bottom of the frame) are shown in green colour,
while the detected track is shown in red colour. 
We use ``baseline+tracks'' in these figure as ``baseline'' method.
Since all the methods use same set of tracks, red box is used to annotate track boxes.
Each method's score has a separate colour, described in the sub-caption.


\par\vfill\par

\clearpage
%
%
\bibliographystyle{splncs04}
\bibliography{ref}